\documentclass{article}

\PassOptionsToPackage{numbers}{natbib}
\usepackage[main,final]{neurips_2026}

\usepackage[utf8]{inputenc}
\usepackage[T1]{fontenc}
\usepackage{microtype}
\usepackage{hyperref}
\usepackage{url}
\usepackage{booktabs}
\usepackage{amsmath}
\usepackage{amssymb}
\usepackage{amsfonts}
\usepackage{mathtools}
\usepackage{graphicx}
\usepackage{subcaption}
\usepackage{xcolor}
\usepackage{xspace}
\usepackage{wrapfig}
\usepackage{caption}

\usepackage{cleveref}

\let\cite\citep
\let\cref\Cref

\title{\method: Generative Dynamic Gaussian Reconstruction from Monocular Video}

\author{%
  {\normalsize\bfseries Liyuan Zhu\textsuperscript{1,2} \quad
  Shengyu Huang\textsuperscript{2} \quad
  Amrita Mazumdar\textsuperscript{2} \quad
  Tianye Li\textsuperscript{2} \quad
  Zan Gojcic\textsuperscript{2}} \\
  {\normalsize\bfseries Gordon Wetzstein\textsuperscript{1} \quad
  Iro Armeni\textsuperscript{1} \quad
  Shalini De Mello\textsuperscript{2} \quad
  Alex Trevithick\textsuperscript{2}} \\[0.25em]
  {\small\normalfont\textsuperscript{1}Stanford University \quad
  \textsuperscript{2}NVIDIA}
  \\[0.35em]
  {\small\normalfont\url{https://research.nvidia.com/labs/amri/projects/world-from-motion/}}
}

\hypersetup{
  pdftitle={World from Motion: Generative Dynamic Gaussian Reconstruction from Monocular Video},
  pdfauthor={Liyuan Zhu; Shengyu Huang; Amrita Mazumdar; Tianye Li; Zan Gojcic; Gordon Wetzstein; Iro Armeni; Shalini De Mello; Alex Trevithick},
  pdfsubject={Generative dynamic 3D Gaussian reconstruction from monocular video}
}

\begin{document}
\newcommand{\methodfull}{World from Motion}
\newcommand{\methodshort}{WfM}
\newcommand{\method}{\methodfull\xspace}
\newcommand{\resulttablesize}{\small}
\newcommand{\metricup}[1]{#1~$\uparrow$}
\newcommand{\metricdown}[1]{#1~$\downarrow$}
\newcommand{\deltagreen}[1]{\textcolor{green!50!black}{\scriptsize($#1$)}}
\newcommand{\deltared}[1]{\textcolor{red!60!black}{\scriptsize($#1$)}}

\newcommand{\scene}{\mathcal{G}}

\def\eg{\emph{e.g.}\xspace} \def\Eg{\emph{E.g.}\xspace}
\def\ie{\emph{i.e.}\xspace} \def\Ie{\emph{I.e.}\xspace}
\def\etc{\emph{etc.}\xspace}
\def\wrt{\emph{w.r.t.}\xspace}
\def\etal{\emph{et al.}\xspace}

\maketitle

\vspace{-1.5em}
\begin{figure}[!ht]
    \centering
    \includegraphics[width=\linewidth]{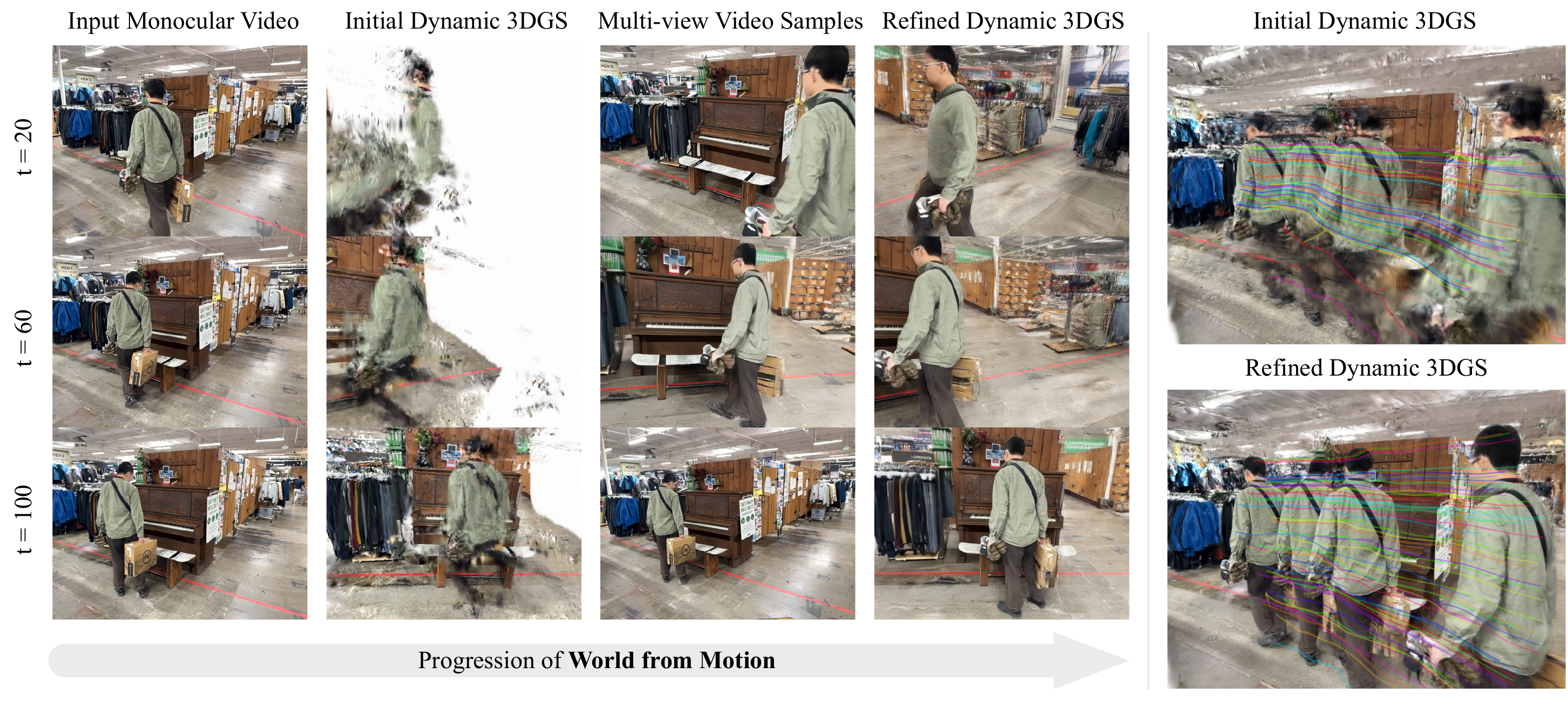}
    \caption{\textbf{\method{} reconstructs a dynamic 3DGS world
    from the camera and scene motion in a single monocular video.}
    From an input video and an initial reconstruction produced by MoSca~\cite{mosca}, our video model generates novel views that are distilled into a refined reconstruction. Our method faithfully recovers observed structure and synthesizes plausible novel-view dynamics, in turn improving the underlying scene motion, as visualized on the right.
    }
    \label{fig:teaser}
\end{figure}

\begin{abstract}
We present World from Motion, a method for generating freely renderable dynamic 3D Gaussian representations from monocular videos. Our approach conditions a video model on dense, pixel-aligned renderings that encode appearance, geometry, and 3D scene motion along both input and target camera trajectories to correct rendering artifacts and fill in missing regions from an initial reconstruction. To train this model, we construct a dataset of aligned multiview video pairs and dynamic 3DGS representations, with simulated artifacts characteristic of monocular reconstruction. At test time, we distill the model's generations, including newly observed regions and motions, back into a single consistent, high-quality dynamic 3DGS, improving both novel-view synthesis and the underlying 3D motion. Our method sets a new state of the art in PSNR, SSIM, and LPIPS on the DyCheck and NVIDIA-LS benchmarks and seamlessly generalizes to in-the-wild videos with large viewpoint changes and dynamic motions.

\end{abstract}

\section{Introduction}
The human visual system excels at inferring the 4D structure of the dynamic world from limited 2D observations. This internal "world model" combines parallax and stereo cues with learned priors, allowing humans to reason about precise geometry and resolve inherent ambiguities in shape, scale, and motion. As a computational realization of this capability, monocular 4D reconstruction enables important applications across robotic simulation, spatial perception, and immersive AR/VR. This process aims to reconstruct a model of the dynamic world's geometry, appearance, and motion for rendering at arbitrary viewpoints and times.

To replicate these human capabilities, classical 3D vision techniques~\cite{slam, sfm, hartley} exploit geometric constraints that are highly effective under sufficient parallax and static scene assumptions. Notably, the outputs of these methods generally improve with more observations. Building on these techniques, the state-of-the-art 4D reconstruction methods~\cite{som, mosca, worldtree} accurately recover well-observed static regions and motion estimates with dynamic 3D Gaussian splatting (3DGS)~\cite{3dgs, luiten2023dynamic_3dgs}. However, they struggle to infer novel views and unobserved regions of dynamic scene elements.

Generative models offer the opposite tradeoff: they can resolve underconstrained regions by filling in missing appearance, geometry, and motion using learned priors. However, while state-of-the-art video generative models~\cite{rcm, veo} can synthesize high-fidelity frames, they often lack the multiview consistency required for precise 4D reconstruction, leading to inconsistent structure when lifted into 3D. Recent hybrid methods~\cite{vidar, cat4d} attempt to combine the strengths of both paradigms, but their results generally lack sufficient multiview consistency for precise dynamic reconstruction (see~\cref{fig:qualitative_dycheck}).
Consequently, existing frameworks are forced to choose between geometric rigor and generative expressivity, failing to capture the fluid yet persistent nature of the dynamic world. This motivates the central question of our work:
\begin{quote}
\itshape
How can a monocular reconstruction method achieve the fidelity of geometry-driven methods where structure is well determined, while simultaneously deferring to generative priors where it is not---especially for scene dynamics?

\end{quote}

While recent work~\cite{liu20243dgs_enhancer,flowr, difix3d, gaussfusion} has made significant progress in leveraging generative priors to enhance \emph{static} scene renderings, extending these successes to \emph{dynamic} environments remains an open challenge.
Dynamic world modeling is fundamentally more difficult, requiring separation of dynamic and static scene elements and estimation of dense 3D scene flow for the dynamic parts.
Precise, generative 4D reconstruction therefore presents two unique challenges: (i) imbuing a generative model with precision and consistency across space-time;
and (ii) lifting the generated samples into a unified, freely renderable 4D representation.

To address these challenges, we propose \emph{\method} (WfM), a 4D-consistent framework that unifies reconstruction and generative modeling. An homage to classical Structure from Motion (SfM)~\cite{sfm}, WfM reconstructs a persistent and traversable snapshot of the world from both camera and scene motion.
Prior methods~\cite{gsdit, cognvs, neoverse} generate only short videos by conditioning on RGB or depth along the target trajectory.
In contrast, WfM generates a high-quality 4D asset. We show how to condition a video model on dense, pixel-aligned renderings of a reconstructed dynamic 3DGS, including explicit 3D scene flow, and distill the generations into the underlying 4D representation. We choose dynamic 3DGS as the conditioning scaffold for two reasons. First, it compactly encodes priors across appearance, geometry, and 3D motion that can be rendered into screen space for dense, aligned video conditioning. Second, its explicit and differentiable form allows it to be consistently updated with novel observations and dynamics.

Put together, our conditional video model sets a new state of the art on the community-standard DyCheck~\cite{dycheck} benchmark in perceptual quality without any further optimization (see~\cref{tab:results}). The dynamic 3DGS optimized using these samples provides even higher-quality results than the generations. In general, our method can handle in-the-wild videos, including large viewpoint changes and 3D motions~(see \cref{fig:teaser}). Concretely, our contributions are as follows:

\begin{enumerate}
    \item We show how to condition a video generative model on dynamic 3DGS  through dense, pixel-aligned appearance, geometry, and 3D-motion buffers generated along both the input \emph{and} target camera trajectories (\cref{sec:4d_generator}).
    \item We design an optimization procedure to distill the generated samples into a unified dynamic 3DGS representation (\cref{subsec:reoptimization}), which enhances motion quality (see~\cref{tab:dycheck_correspondence}) and improves with more generations (see~\cref{tab:iphone_multiview_camera_count}).
    \item We construct a carefully designed dataset of paired dynamic 3D Gaussian representations and ground-truth multiview videos, simulating the artifacts of monocular capture (\cref{sec:training_artifacts}).

\end{enumerate}

\section{Related Work}

\vspace{-0.5em}
\paragraph{4D reconstruction from monocular video.}

Casual monocular videos usually capture the dynamic nature of the real world.
Yet classical methods for camera pose and depth estimation typically assume a static capture~\cite{sfm, slam, slam2}. Early non-rigid shape-from-motion methods relaxed this assumption using low-rank trajectory priors~\cite{bregler2000recovering}. While recent systems improve robustness through learned optimization~\cite{teed2021droid}, long-term tracks~\cite{li2025megasam}, or dense point maps~\cite{huang2025vipe, monst3r, dust3r, wang2025vggt,xu2025depth, zhou2025page, karhade2026any4d}, inferring a unified representation including unobserved content remains challenging.

Beyond camera and depth estimation, monocular 4D reconstruction must infer geometry and scene flow in a unified frame.
NeRF-based methods~\cite{rodynrf, nsff, dynpoint, hypernerf, dnerf, tineuvox, nerfies} briefly dominated dynamic view synthesis before dynamic 3DGS optimization-based methods~\cite{som, mosca, worldtree, marbles} became state-of-the-art. These methods leverage long-range motion supervision but struggle to infer unseen dynamics and content. CAT4D~\cite{cat4d} and ViDAR~\cite{vidar} incorporate generative models into this optimization but require SDEdit-style~\cite{meng2022sdedit} noising and denoising, resulting in inconsistent samples and blurry output. In parallel, feedforward approaches aim to predict 4D structure directly: 4DGT~\cite{4dgt}, MoVieS~\cite{movies}, and Lyra~\cite{lyra} directly predict dynamic 3D Gaussian parameters. While these feedforward methods improve speed, they retain a limited context window and therefore lag behind optimization-based reconstruction in geometric fidelity over longer videos.

\vspace{-0.5em}
\paragraph{3D-Conditioned Visual Generative Models.}

Early work in 3D reconstruction with diffusion priors regularized synthesis from text or sparse images~\cite{dreamfusion, vsd, sparsefusion, dreamcraft3d, wu2024reconfusion, liu20243dgs_enhancer}. A line of image-space "fixers" seeks to enhance 3DGS via conditioning on imperfect renderings. Difix3D~\cite{difix3d} repurposes a single-step image generator for rendering enhancement. FlowR~\cite{flowr} employs a multiview flow-matching model to fix renderings from multiple viewpoints. GaussFusion~\cite{gaussfusion} proposes to condition video diffusion models on 3DGS for enhanced 3D reconstruction. However, although these methods perform well in static settings, they do not target the more challenging 4D regime.

A second paradigm, camera-controlled video generation, encodes pose information to synthesize sequences along specified trajectories~\cite{he2024cameractrl, rcm}, with PlenopticDreamer~\cite{PlenopticDreamer} introducing camera-memory banks for long-term consistency. While visually plausible, these models typically prioritize 2D video quality over persistent 4D assets.
Finally, recent work has sought to anchor generation in explicit 3D/4D representations~\cite{trajcrafter,yu2024viewcrafter,neoverse}. Gen3C~\cite{gen3c} conditions on a point-based representation; VMem~\cite{li2025vmem} retrieves past observations through surfel-indexed memory and SPMem~\cite{spmem} combines 3D structure with 2D memory.  Vivid4D~\cite{huang2025vivid4d} employs depth-guided inpainting. Concurrent work Vista4D~\cite{vista4d} conditions on dynamic point clouds; however, due to this weak conditioning, its results fall short of the precision required for high-fidelity 4D reconstruction (see~\cref{fig:qualitative_dycheck}). GS-DiT~\cite{gsdit}, CogNVS~\cite{cognvs}, and NeoVerse~\cite{neoverse} condition video generators on RGB (and depth) renderings of a dynamic reconstruction along the target trajectory, but output per-trajectory videos without updating the underlying 4D representation. In contrast, our approach conditions a video generator on geometry, appearance, and explicit 3D scene-flow cues from dynamic 3DGS renderings along both reference and target views, and distills the generations back into that same representation. This process allows us to generate not only videos, but renderable 4D worlds with improved 3D motion.

\begin{figure}[t]
    \centering
    \begingroup
    \newcommand{\wildmethod}[1]{\makebox[0.15783\linewidth][c]{\scriptsize #1}}
    \wildmethod{Input Video}\hspace{0.00247\linewidth}%
    \wildmethod{Initial Recon.}\hspace{0.00247\linewidth}%
    \wildmethod{Ours (Distill)}\hspace{0.04192\linewidth}%
    \wildmethod{Input Video}\hspace{0.00247\linewidth}%
    \wildmethod{Initial Recon.}\hspace{0.00247\linewidth}%
    \wildmethod{Ours (Distill)}%
    \par\vspace{0.2em}
    \includegraphics[width=\linewidth]{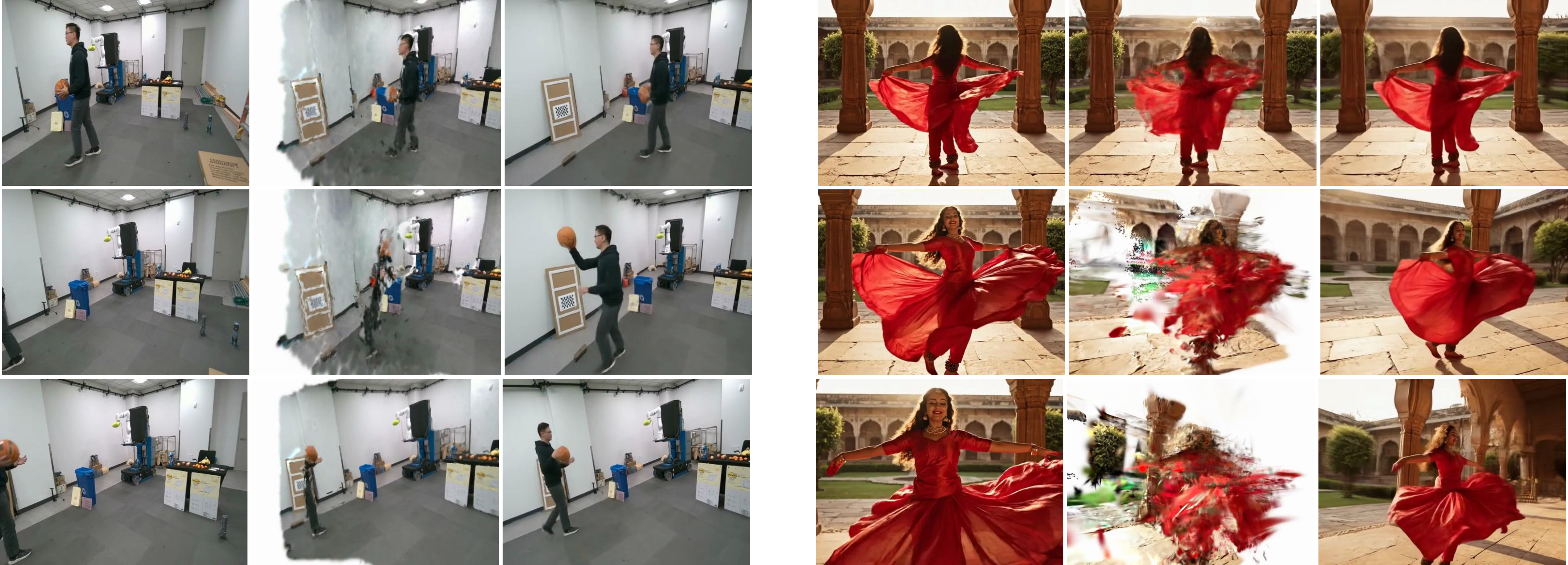}
    \endgroup
    \caption{\textbf{Qualitative results on challenging in-the-wild videos}. Compared to the input reconstruction~\cite{mosca}, our method performs visual outpainting, fixes degraded dynamics, and infers out-of-frustum dynamics while respecting the static region.}
    \label{fig:wild}
\end{figure}

\section{Method}

\vspace{-0.5em}
\paragraph{Overview.}

Given a monocular input video $\mathbf{V}=\{\mathbf{I}_t\}_{t=1}^{N}$ with
$\mathbf{I}_t \in \mathbb{R}^{H \times W \times 3}$, our goal is to recover a dynamic 3DGS representation $\mathcal{G}$ that can produce novel-view renderings from arbitrary viewpoints at arbitrary times. Our approach begins by estimating an initial dynamic 3DGS representation $\mathcal{G}_0$
with an existing method (second part of \cref{subsec:prelim}). We then proceed in two stages. First, we show how to generate virtual observations of the evolving scene from fixed cameras by conditioning a video generator on the initial reconstruction's rendering (\cref{sec:4d_generator}). Then, we show how to distill the generated, high-quality videos back into the 4D representation through re-optimization (\cref{subsec:reoptimization}).

\subsection{Preliminaries}
\label{subsec:prelim}

\vspace{-0.5em}
\paragraph{Video generation model.}
\label{prelim:vdm}
Latent video diffusion models~\cite{Latent_video_diffsion} generate videos in the
latent space of a variational autoencoder (VAE). Given a video
$\mathbf{V}=\{\mathbf{I}_t\}_{t=1}^{N}$, the VAE encoder $\mathcal{E}$ maps it to
a latent tensor $\mathbf{z}_0=\mathcal{E}(\mathbf{V})$, and a decoder maps sampled latents back to RGB frames. These models are commonly trained with flow matching~\cite{flowmatching_loss}. Given Gaussian noise $\boldsymbol{\epsilon}\sim\mathcal{N}(\mathbf{0},\mathbf{I})$
and an interpolation
$\mathbf{z}_{\tau}=(1-\tau)\boldsymbol{\epsilon}+\tau\mathbf{z}_0$ for
$\tau\in[0,1]$, the model learns a velocity field $\mathbf{v}_{\theta}$ that predicts the
direction from noise to the data latent by minimizing
$\mathcal{L}_{\mathrm{fm}}=\|\mathbf{v}_{\theta}(\mathbf{z}_{\tau},\tau;\mathbf{c})-
(\mathbf{z}_0-\boldsymbol{\epsilon})\|_2^2$, conditioned
on text, images, or other controls $\mathbf{c}$. Modern video generators usually
adopt a Diffusion Transformer (DiT) backbone~\cite{dit} to parameterize this
velocity field.

\vspace{-0.5em}
\paragraph{Monocular 4D reconstruction.}
\label{prelim:4d_recon}
Given the input monocular video $\mathbf{V}$, we first obtain an initial dynamic 3DGS representation by running an off-the-shelf monocular 4D reconstruction method~\cite{som,mosca,worldtree}. These methods start by estimating depth~$\mathcal{D}$ and camera parameters
$\mathcal{P}^{\mathrm{in}}=\{(\mathbf{K}_t,\mathbf{T}_t)\}_{t=1}^{N}$, where
$\mathbf{K}_t$ denotes intrinsics and $\mathbf{T}_t \in SE(3)$ denotes the
camera pose~\cite{xu2025depth, li2025megasam}. They also employ priors to predict dynamic masks~\cite{sam3, mosca}, and long-term tracks~\cite{doersch2023tapir,xiao2025spatialtrackerv2}. The extracted camera poses and depth maps are then used to lift 2D tracks to 3D to initialize a low-rank motion representation, \eg, motion bases~\cite{som}, motion scaffolds~\cite{mosca}, and temporal tree structures~\cite{worldtree}. The dynamic scene $\mathcal{G}$ is represented by static and dynamic 3D Gaussians with
$M$ Gaussian primitives and the underlying motion representation that maps canonical Gaussian attributes to their state at time $t$.

At any given time $t$, the state of each Gaussian is evaluated as $g_i(t)=(\boldsymbol{\mu}_i(t),\mathbf{R}_i(t),\mathbf{s}_i,\alpha_i,\mathbf{c}_i)$. In this formulation, the 3D center $\boldsymbol{\mu}_i(t) \in \mathbb{R}^{3}$ and orientation $\mathbf{R}_i(t) \in SO(3)$ are time-dependent, mapped from canonical attributes to their current state by the motion model. The remaining properties including scale $\mathbf{s}_i \in \mathbb{R}^{3}$, opacity $\alpha_i \in \mathbb{R}$, and view-dependent color coefficients $\mathbf{c}_i$, are stored as constant, per-Gaussian attributes. Once evaluated at time $t$, the standard rasterization pipeline~\cite{3dgs} projects each 3D Gaussian to a 2D splat and alpha-composites the visible splats in depth order from a camera viewpoint $(\mathbf{K},\mathbf{T})$ to form the rendered image $\hat{\mathbf{I}}_t = \mathcal{R}_{\mathrm{rgb}}(\mathcal{G},t,\mathbf{K},\mathbf{T})$. Along with the photometric loss, the extracted priors supervise a depth loss $\mathcal{L}_{\mathrm{depth}}$ and a 3D track loss $\mathcal{L}_{\mathrm{track}}$, and an as-rigid-as-possible term $\mathcal{L}_{\mathrm{arap}}$ regularizes the motion. We refer readers to \cite{som,mosca,worldtree} for details.

Off-the-shelf methods provide an initial dynamic 3DGS~$\mathcal{G}_0$ which exhibits
floaters, missing regions, blurry texture, and incorrect dynamics due to heavy regularization to compensate for the ill-posedness of the problem. Nevertheless, it still provides a persistent 4D scene scaffold that can be rendered from novel camera trajectories.
We now describe how to repurpose a video generator to condition on this 4D scene representation for view synthesis across space and time.

\begin{figure}[t]
    \centering
    \includegraphics[width= \linewidth]{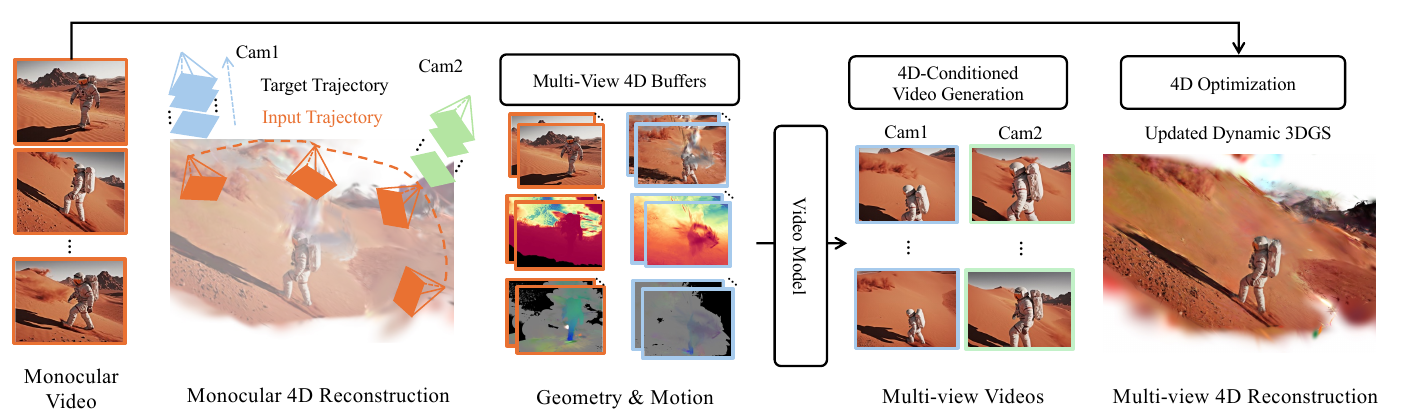}
    \vspace{-1em}
    \caption{
        \textbf{Overview of 4D reconstruction with \method}. Given a monocular video, our method first generates an initial 4D reconstruction of the input video. Along the target trajectory, our method then renders the corresponding appearance, geometry and motion to condition a video generative model. The generated samples are used to create a higher-quality 4D reconstruction.
    }
    \label{fig:overview}
\end{figure}

\subsection{Dynamic 3DGS-conditioned Target Video Generation}
\label{sec:4d_generator}

We choose to condition our video generator on pixel-aligned multi-modal renderings (4D buffers) of a dynamic 3DGS rather than on point clouds~\cite{gen3c,vista4d} or posed images~\cite{rcm,trajcrafter} because dynamic 3DGS provides a persistent, continuous, and differentiable 4D scene scaffold. While recent works~\cite{gaussfusion, flowr, difix3d} have demonstrated the effectiveness of conditioning generative models on 3DGS renderings for static 3D reconstruction, extending this paradigm to dynamic 4D settings is nontrivial. Monocular 4D reconstruction is inherently more ambiguous, making rendered buffers from novel target views less reliable, especially for dynamic regions.

\vspace{-0.5em}
\paragraph{Rendering 4D buffers.}
Due to the ambiguities of 4D reconstruction, we condition our video model on as much scene information as possible. Prior work in the static regime conditions video generative models only on information from the target trajectory~\cite{gaussfusion}, which can fail to anchor generated content to the observed input. To address this, we explicitly condition on 4D buffers rendered along both the input trajectory $\mathcal{P}^{\mathrm{in}}$ and the target trajectory $\mathcal{P}^{\mathrm{tgt}}$, providing stronger and more stable guidance for dynamic scene synthesis.

Concretely, given the initial representation $\mathcal{G}_0$, we render per-frame conditioning signals for any camera viewpoint $(\mathbf{K}_t, \mathbf{T}_t)$ at time $t$ by evaluating the dynamic Gaussian state $\mathcal{G}_0(t)$. Standard rasterization produces the RGB image $\hat{\mathbf{I}}_t$ along with spatial G-buffers including opacity $\mathbf{A}_t$, depth $\mathbf{D}_t$, and surface normals $\mathbf{N}_t$, which provide dense geometric and appearance cues.
To further encode scene dynamics, we augment these signals with 3D scene flow defined in the global coordinate system of $\mathcal{G}_0$. Specifically, we compute per-Gaussian displacements $\mathbf{u}_i(t)=\boldsymbol{\mu}_i(t+1)-\boldsymbol{\mu}_i(t)$ and rasterize them into a per-pixel flow map $\mathbf{F}_t = \mathcal{R}_{\mathrm{flow}}(\mathcal{G}_0,t,\mathbf{K}_t,\mathbf{T}_t)$. Each pixel thus stores a full 3D displacement in world coordinates rather than a 2D optical flow.
We aggregate these signals into a per-frame conditioning bundle $\mathbf{b}_t = [\hat{\mathbf{I}}_t, \mathbf{A}_t, \mathbf{D}_t, \mathbf{N}_t, \mathbf{F}_t]$, and render sequences $\mathbf{B}^{\mathrm{in}}=\{\mathbf{b}_t^{\mathrm{in}}\}_{t=1}^N$ and $\mathbf{B}^{\mathrm{tgt}}=\{\mathbf{b}_t^{\mathrm{tgt}}\}_{t=1}^N$ along the input and target trajectories, respectively.

\vspace{-0.5em}
\paragraph{Model architecture.}
Our goal is to enable the generator to preserve reliable regions of the initial 4D reconstruction while correcting its failure modes, including inaccurate dynamics and missing content, in a temporally and multiview-consistent manner.
Given the buffer sequences $\mathbf{B}^{\mathrm{in}}$ and $\mathbf{B}^{\mathrm{tgt}}$, we first encode each modality independently using the video VAE encoder $\mathcal{E}$. The encoded modalities are concatenated along the channel dimension to form per-frame 4D tokens, and then concatenated along the temporal dimension across input and target trajectories, yielding the full conditioning sequence $\mathbf{z}_{\mathbf{B}}$. This sequence is processed by a VACE-style~\cite{vace} DiT adapter (the \emph{4D buffer adapter}), where self-attention over the 4D tokens produces residual updates that are injected into each layer of the base DiT.
The base DiT takes in the encoded input and target videos separately, applies noise only to the target latents, and concatenates clean input latents with noisy target latents along the temporal dimension before the DiT blocks. In addition to the 4D buffer adapter, we incorporate explicit camera conditioning by encoding each camera pose $\tilde{\mathbf{T}}$ into a frame-level embedding, which is broadcast to all spatial tokens before self-attention, providing geometric guidance for both input and target sequences.

We train the model using the flow-matching loss defined in \cref{prelim:vdm}, applied only to the target latents. \cref{fig:network_architecture} in the supplement illustrates the full architecture. Since the conditioning is derived from rendered 4D buffers rather than representation-specific parameters, the model naturally generalizes across different underlying 4D representations.

\paragraph{Reconstruction guidance.}
We observe that 4D buffer-based guidance improves alignment between generated samples and the underlying 4D scene representation. To enable this, we introduce a classifier-free-style dropout strategy on the 4D buffer tokens. During training, we randomly replace the encoded 4D-buffer tokens $\mathbf{z}_{\mathbf{B}}$ (for both input and target trajectories) with a learned null embedding $\bar{\mathbf{z}}_{\mathbf{B}}$. Importantly, the base conditioning from the input video and camera poses is always retained, so the resulting “unconditional” branch corresponds to a ReCamMaster-like model~\cite{rcm} without 4D buffer guidance.

At inference time, we evaluate the model twice, with and without the 4D buffer condition, producing velocities $\mathbf{v}_{\theta,c}$ and $\mathbf{v}_{\theta,u}$, respectively. We then apply a guidance update analogous to classifier-free guidance to steer the generation toward the reconstruction-conditioned prediction:
\begin{equation}
\label{eq:geometry_guidance}
\mathbf{v}_{\theta,\mathrm{geo}} =
\mathbf{v}_{\theta,u} + \gamma(\tau)\big(\mathbf{v}_{\theta,c} - \mathbf{v}_{\theta,u}\big),
\end{equation}
where $\gamma(\tau)$ is a timestep-dependent guidance scale. In practice, we also support more general guidance operators such as APG~\cite{apg}. This guidance improves geometric consistency by explicitly controlling the influence of the 4D reconstruction prior during generation.

\subsection{Refining the 4D Scene Representation}
\label{subsec:reoptimization}

Our goal is to leverage the trained video generator to synthesize novel observations of the 4D scene $\mathcal{G}_0$ and use them to refine the underlying representation. We sample a set of target viewpoints and generate temporally consistent videos from these cameras. These synthesized observations are then used together with the input video to re-optimize $\mathcal{G}_0$.

\vspace{-0.5em}
\paragraph{Sampling virtual cameras.}
To ensure good spatial coverage, we select $J$ target viewpoints via farthest-point sampling over the
input camera trajectory and keep them fixed over time. For each sampled
trajectory $\mathcal{P}^{\mathrm{tgt},j}$, we generate a video
$\mathbf{V}^{\mathrm{tgt},j}=\{\mathbf{I}^{\mathrm{tgt},j}_{t}\}_{t=1}^{N}$
using the procedure described in \cref{sec:4d_generator}. Virtual cameras are sampled from the input trajectory only; test trajectories are never used for conditioning or supervision. For long sequences, generation is performed in a temporal sliding-window manner.

We empirically find that this simple sampling strategy is sufficient to produce target videos with sufficient spatial coverage and strong spatio-temporal consistency. This is enabled by our 4D-conditioned generator, which maintains coherence across viewpoints without requiring complex cross-view coordination. In contrast, prior work such as CAT4D~\cite{cat4d} relies on more involved multi-view and temporal sampling procedures to achieve consistent generation.

\vspace{-0.5em}
\paragraph{Reoptimization.}
We refine the initial dynamic 3D Gaussian
representation $\mathcal{G}_0$ by re-optimizing the initialized model with supervision from both the
input video $\mathbf{V}$ along $\mathcal{P}^{\mathrm{in}}$ and the generated videos
$\{\mathbf{V}^{tgt,j}\}_{j=1}^{J}$ along $\{\mathcal{P}^{tgt,j}\}_{j=1}^{J}$.
During initialization, we jointly estimate the depth of input and generated videos together with DAv3~\cite{xu2025depth}, then backproject the depth maps of the generated views to initialize additional Gaussians to fill in newly-generated content in the scene.
We jointly optimize the Gaussian variables, including 3D centers, rotations, scales, opacities, and motion parameters. We find re-optimization of the motion representation critical because it allows our synthesized samples to correct potentially inaccurate estimates from the initial 3D tracks of the input view.
Overall, we optimize the following loss:
\begin{equation}
\label{eq:reopt_objective}
\mathcal{L}_{\mathrm{reopt}} =
\mathcal{L}_{\mathrm{in}}
+
\lambda_{\mathrm{tgt}}
\mathcal{L}_{\mathrm{tgt}}
+
\lambda_{\mathrm{motion}}
\mathcal{L}_{\mathrm{motion}}
+
\lambda_{\mathrm{depth}}
\mathcal{L}_{\mathrm{depth}}
+
\lambda_{\mathrm{track}}
\mathcal{L}_{\mathrm{track}}
+
\lambda_{\mathrm{arap}}
\mathcal{L}_{\mathrm{arap}}
\end{equation}
where each $\lambda\in\mathbb{R}$ controls the weight of its respective loss component.
$\mathcal{L}_{\mathrm{depth}}$ and $\mathcal{L}_{\mathrm{track}}$ are supervised by the estimated depth and 3D tracks extracted in the initial reconstruction (\cref{prelim:4d_recon}).
The input term $\mathcal{L}_{\mathrm{in}}$ keeps the representation faithful to the observed input video,
while the generated-view term $\mathcal{L}_{\mathrm{tgt}}$ uses generated videos as additional supervision.
Specifically, $\mathcal{L}_{\mathrm{in}}$ sums a per-frame appearance loss $\ell_{\mathrm{app}}$
between renderings from $\mathcal{G}_{0}$ and the input video
$\mathbf{V}$ along $\mathcal{P}^{\mathrm{in}}$, and $\mathcal{L}_{\mathrm{tgt}}$ sums
$\ell_{\mathrm{app}}$ over all generated videos $\{\mathbf{V}^{tgt,j}\}_{j=1}^{J}$
along their sampled camera trajectories $\{\mathcal{P}^{tgt,j}\}_{j=1}^{J}$. This appearance loss consists of photometric and perceptual terms:
\begin{equation}
\label{eq:appearance_loss}
\ell_{\mathrm{app}}(\hat{\mathbf{I}},\mathbf{I})
=
\|\hat{\mathbf{I}}-\mathbf{I}\|_1
+
\lambda_{\mathrm{ssim}}\mathcal{L}_{\mathrm{SSIM}}(\hat{\mathbf{I}},\mathbf{I})
+
\lambda_{\mathrm{lpips}}\mathcal{L}_{\mathrm{LPIPS}}(\hat{\mathbf{I}},\mathbf{I}).
\end{equation}
$\mathcal{L}_{\mathrm{in}}$ and $\mathcal{L}_{\mathrm{tgt}}$ supervise both
the underlying motion representation and the appearance parameters of the
dynamic Gaussians, while $\mathcal{L}_{\mathrm{arap}}$ is defined in~\cite{mosca}.

\begin{figure}[t]
    \centering
    \begingroup
    \newcommand{\figmethod}[1]{\makebox[0.142857\linewidth][c]{\scriptsize #1}}
    \figmethod{Vista4D\textsuperscript{$\dagger$}~\cite{vista4d}}%
    \figmethod{CAT4D~\cite{cat4d}}%
    \figmethod{ViDAR~\cite{vidar}}%
    \figmethod{WorldTree~\cite{worldtree}}%
    \figmethod{Ours (distill)}%
    \figmethod{Ours (sample)}%
    \figmethod{Ground Truth}%
    \par\vspace{0.2em}
    \includegraphics[width=\linewidth,trim={0 491.3pt 0 0},clip]{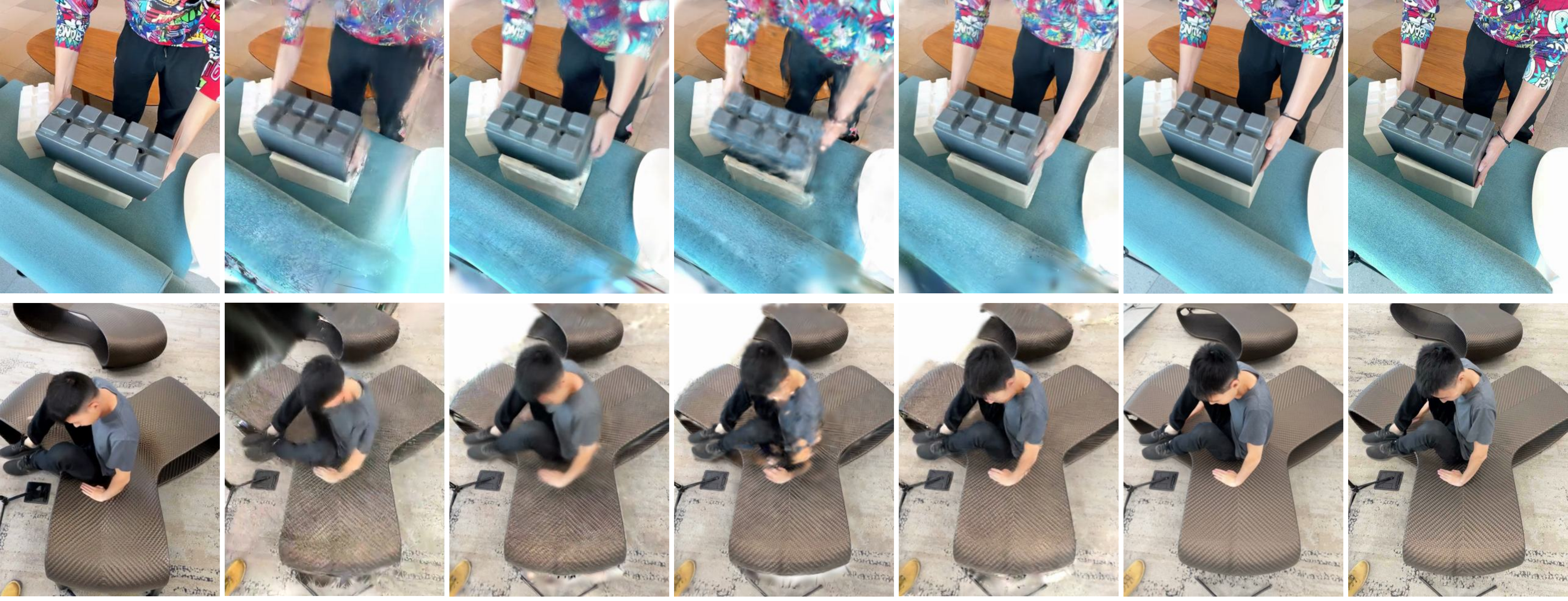}
    \endgroup
    \caption{\textbf{Qualitative comparison on DyCheck~\cite{dycheck}.} Each row compares the predicted test view from each method with the ground-truth frame.
Our method produces sharper details and consistency.}
    \label{fig:qualitative_dycheck}
\end{figure}

\section{Generating Training Data from Multiview Videos}

\label{sec:training_artifacts}
\vspace{-0.5em}
\paragraph{Training data.}
We train our model on synthetic multiview dynamic scenes from the MultiCamVideo dataset~\cite{rcm}, which provides posed and time-synchronized multiview videos. For each scene, we randomly select one video as input and another as target to form training pairs. We reconstruct an initial dynamic 3DGS representation from the input video using an off-the-shelf monocular 4D reconstruction method, such as Shape of Motion~\cite{som} or WorldTree~\cite{worldtree}, and render it along both input and target trajectories to obtain the 4D conditioning buffers.

\paragraph{Joint depth alignment.} Geometric alignment between rendered conditioning and ground-truth video is critical; scale ambiguity in monocular reconstruction often leads to inconsistencies that cause the generator to ignore the conditioning signal. We resolve this by enforcing global scale consistency through joint depth prediction across views. Specifically, we utilize anchor frames with large baselines to provide multi-view constraints, coupling otherwise independent monocular estimates within each prediction window. By aligning the resulting trajectory to the ground-truth and propagating the recovered metric scale via overlapping frames, we produce consistently scaled depth that ensures the conditioning remains spatially compatible with the target video.

\vspace{-0.5em}
\section{Experiments}

\vspace{-0.5em}
\paragraph{Baseline methods.}
We compare against non-generative, generative, and hybrid approaches for 4D reconstruction. Non-generative baselines include Shape of Motion~\cite{som}, MoSca~\cite{mosca}, and WorldTree~\cite{worldtree}. Hybrid methods include CAT4D~\cite{cat4d} and ViDAR~\cite{vidar}. We further consider generative approaches that do not optimize an explicit dynamic 3DGS representation, including ReCamMaster~\cite{rcm} and concurrent work Vista4D~\cite{vista4d}. For our method, we report the sampled videos (\textit{Ours (sample)}) and distilled dynamic 3DGS representation (\textit{Ours (distill)}). \textit{Ours (sample)} queries the generator directly on the test trajectory to benchmark generation against other generative methods on equal footing; these samples play no role in distillation. \textit{Ours (distill)} is supervised only by the input video and by virtual cameras sampled from the input trajectory, so test camera trajectories are never used when generating virtual views, and all distilled results are evaluated on held-out views.

\vspace{-0.5em}
\paragraph{Evaluation datasets.}
We evaluate our method on three benchmarks. First, we report results on the \textit{DyCheck} benchmark~\cite{dycheck}, a standard dataset for dynamic scene reconstruction. Second, we use a held-out validation set from \textit{MultiCamVideo}~\cite{rcm}, consisting of 5 synthetic multiview videos of dynamic human scenes. Third, we evaluate on the real-world NVIDIA-LS dataset~\cite{nvidials} following the protocol of WorldTree~\cite{worldtree}. In addition, we present qualitative results on challenging in-the-wild videos, including both real-world footage and videos synthesized by a generative model~\cite{veo}.

\begin{table*}[t]
    \centering
    \resulttablesize
    \setlength{\tabcolsep}{0.9pt}
    \caption{\textbf{Quantitative results on the DyCheck dataset.}
    We evaluate under the static-camera protocol using three regions (\textit{Full, Valid, Covisible}), and additionally report results with per-frame pose optimization for comparison with prior work.
    All metrics are computed over the corresponding evaluation masks, except for the full (unmasked) evaluation.%
    }
    \label{tab:results}
    \resizebox{\textwidth}{!}{%
    \begin{tabular}{@{}l*{12}{c}@{}}
        \toprule
        & \multicolumn{9}{c}{Static-camera evaluation}
        & \multicolumn{3}{c}{Per-frame pose opt.} \\
        \cmidrule(lr){2-10}
        \cmidrule(lr){11-13}
        & \multicolumn{3}{c}{Full}
        & \multicolumn{3}{c}{Valid}
        & \multicolumn{3}{c}{Covisible}
        & \multicolumn{3}{c}{Covisible} \\
        \cmidrule(lr){2-4}
        \cmidrule(lr){5-7}
        \cmidrule(lr){8-10}
        \cmidrule(lr){11-13}
        Method
        & PSNR $\uparrow$ & SSIM $\uparrow$ & LPIPS $\downarrow$
        & mPSNR $\uparrow$ & mSSIM $\uparrow$ & mLPIPS $\downarrow$
        & mPSNR $\uparrow$ & mSSIM $\uparrow$ & mLPIPS $\downarrow$
        & mPSNR $\uparrow$ & mSSIM $\uparrow$ & mLPIPS $\downarrow$ \\
        \midrule
        Shape-of-Motion~\cite{som} & 16.89 & 0.477 & 0.340 & 17.27 & 0.569 & 0.286 & 17.32 & 0.598 & 0.296 & -- & -- & -- \\
        MoSca (MS)~\cite{mosca} & 16.60 & 0.553 & 0.362 & 17.76 & 0.644 & 0.303 & 18.69 & 0.696 & 0.272 & 19.32 & 0.706 & 0.264 \\
        WorldTree (WT)~\cite{worldtree} & 17.08 & 0.574 & 0.332 & 18.38 & 0.662 & 0.278 & 19.28 & 0.714 & 0.246 & 19.75 & 0.728 & 0.240 \\
        ViDAR~\cite{vidar} & 17.22 & 0.570 & 0.315 & 18.64 & 0.661 & 0.254 & 19.44 & 0.711 & 0.224 & 19.69 & 0.713 & 0.223 \\
        \midrule
        MS + Ours (sample) & 18.64 & 0.556 & \textbf{0.257} & 19.01 & 0.646 & 0.229 & 19.45 & 0.694 & 0.214 & -- & -- & -- \\
        MS + Ours (distill) & 17.30 & 0.566 & 0.268 & 18.73 & 0.657 & \textbf{0.206} & 19.63 & 0.707 & \textbf{0.184} & 20.26 & 0.725 & \textbf{0.176} \\
        WT + Ours (sample) & \textbf{18.79} & 0.563 & 0.259 & \textbf{19.17} & 0.652 & 0.230 & 19.63 & 0.701 & 0.214 & -- & -- & -- \\
        WT + Ours (distill) & 17.57 & \textbf{0.583} & 0.271 & 19.10 & \textbf{0.672} & 0.211 & \textbf{20.06} & \textbf{0.721} & 0.187 & \textbf{20.38} & \textbf{0.730} & 0.184 \\
        \bottomrule
    \end{tabular}%
    }
\end{table*}

\begin{table}[t]
    \centering

\begin{minipage}[t]{0.52\columnwidth}
    \vspace{0pt}
    \centering
    \resulttablesize
    \captionof{table}{\textbf{3DGS refinement with varying numbers of virtual cameras.}}
    \label{tab:iphone_multiview_camera_count}

    \vspace{0.35em}
    \setlength{\tabcolsep}{2pt}
    \resizebox{0.9\linewidth}{!}{%
    \begin{tabular}{c|cccccc}
        \toprule
        \# virtual cams & 0 & 1 & 2 & 4 & 8 & 16 \\
        \midrule
        mPSNR $\uparrow$ & 18.69 & 19.29 & 19.39 & 19.49 & \textbf{19.78} & 19.63 \\
        \bottomrule
    \end{tabular}%
    }
\end{minipage}\hfill
\begin{minipage}[t]{0.44\columnwidth}
    \vspace{0pt}
    \centering
    \resulttablesize
    \captionof{table}{\textbf{DyCheck 3D track evaluation}. PCK@0.05 $\uparrow$ is reported.}
    \label{tab:dycheck_correspondence}

    \vspace{0.35em}
    \setlength{\tabcolsep}{2pt}
    \resizebox{\linewidth}{!}{%
    \begin{tabular}{@{}cccc@{}}
        \toprule
        CoTracker~\cite{karaev23cotracker} & BootsTAPIR~\cite{doersch2024bootstap} & MoSca~\cite{mosca} & Ours \\
        \midrule
        0.803 & 0.779 & 0.824 & \textbf{0.862} \\
        \bottomrule
    \end{tabular}%
    }
\end{minipage}
\end{table}

\vspace{-0.5em}
\paragraph{DyCheck benchmark~\cite{dycheck}.}

We present quantitative results in~\cref{tab:results} using two reconstruction backbones, WorldTree~\cite{worldtree} and MoSca~\cite{mosca}. The DyCheck benchmark assumes static test cameras over time. However, several prior works~\cite{mosca,worldtree,vidar} perform per-frame pose re-optimization during evaluation, which alters the camera trajectory and deviates from this assumption. To ensure a fair comparison, we recalibrate these methods to use fixed camera poses across time. For completeness, we also report results under the original protocol with per-frame pose optimization (see \cref{sec:pose_optimization} for more analysis). Under the static-camera setting, we evaluate performance over three regions: (1) \textit{covisible}, which includes only pixels visible in both reference and test views at each timestep~\cite{dycheck}; (2) \textit{valid}, defined as the union of covisible regions across all timesteps; and (3) the \textit{full} image.

Across all settings, our method consistently outperforms prior approaches. The video generator alone achieves higher PSNR and lower LPIPS than all baselines, including when applied to out-of-distribution reconstructions from MoSca. After distillation, the reconstructed 4D representation further improves accuracy, yielding the best performance across all metrics and regions. We further compare against generative approaches in~\cref{tab:generation_results}, where our method (sample) outperforms all baselines, including concurrent Vista4D~\cite{vista4d}, demonstrating the benefit of conditioning on a structured and geometrically aligned 4D representation (\ie WorldTree). As seen in~\cref{tab:dycheck_correspondence}, compared to the input MoSca~\cite{mosca}, the photometric constraints from our samples improve the underlying 3D tracks. Under the protocols of CogNVS~\cite{cognvs} and NeoVerse~\cite{neoverse}, which differ from~\cref{tab:results} in resolution, masks, and pose handling, our method outperforms both by large margins on all metrics (\cref{tab:protocol_comparison}).

Qualitative comparisons in~\cref{fig:qualitative_dycheck} show that our method produces sharper, more coherent results with improved geometric alignment. Compared to generative methods such as Vista4D~\cite{vista4d} and CAT4D~\cite{cat4d}, our results better preserve structure while reducing temporal inconsistencies and geometric drift. Compared to reconstruction-based methods such as WorldTree~\cite{worldtree} and ViDAR~\cite{vidar}, our method significantly improves fidelity, reducing blurring and high-frequency noise.

\vspace{-0.5em}
\vspace{-0.5em}
\paragraph{NVIDIA-LS~\cite{nvidials} and MultiCamVideo~\cite{rcm} benchmarks.}

On the real-world NVIDIA-LS dataset, we sample 4 virtual cameras and outperform all baselines on all metrics (\cref{tab:nvidia_ls}). On MultiCamVideo, \cref{tab:dataset_results} presents the quantitative comparison with baselines, where our method achieves the best performance across all metrics. The results show that our framework effectively recovers disoccluded regions and stabilizes ambiguous motion in monocular inputs. We refer readers to the supplement and video for qualitative comparisons.

\begin{table}[t]
    \centering
    \footnotesize
    \setlength{\tabcolsep}{4pt}
    \begin{minipage}[t]{0.485\linewidth}
        \vspace{0pt}
        \centering
        \captionof{table}{\textbf{Comparison against generative methods on DyCheck.}}
        \label{tab:generation_results}
        \vspace{0.2em}
        \begin{tabular}{@{}lccc@{}}
            \toprule
            Method & \metricup{mPSNR} & \metricup{mSSIM} & \metricdown{mLPIPS} \\
            \midrule
            CAT4D~\cite{cat4d} & 18.24 & 0.666 & 0.227 \\
            Ours & \textbf{19.89} & \textbf{0.715} & \textbf{0.197} \\
            \midrule
            ReCamMaster~\cite{rcm} & 10.96 & 0.262 & 0.755 \\
            TrajectoryCrafter~\cite{trajcrafter} & 13.06 & 0.320 & 0.656 \\
            GEN3C~\cite{gen3c} & 12.06 & 0.260 & 0.679 \\
            Vista4D~\cite{vista4d} & 14.14 & 0.310 & 0.514 \\
            Ours (eval at full-res) & \textbf{18.45} & \textbf{0.635} & \textbf{0.362} \\
            \bottomrule
        \end{tabular}
    \end{minipage}\hfill
    \begin{minipage}[t]{0.485\linewidth}
        \vspace{0pt}
        \centering
        \captionof{table}{\textbf{Comparison under the DyCheck protocols of CogNVS~\cite{cognvs} (top) and NeoVerse~\cite{neoverse} (bottom).} $^\dagger$As reported in~\cite{neoverse}.}
        \label{tab:protocol_comparison}
        \vspace{0.2em}
        \begin{tabular}{@{}lccc@{}}
            \toprule
            Method & \metricup{PSNR} & \metricup{SSIM} & \metricdown{LPIPS} \\
            \midrule
            TrajectoryCrafter~\cite{trajcrafter} & 12.94 & 0.279 & 0.590 \\
            CogNVS~\cite{cognvs} & 15.36 & 0.356 & 0.426 \\
            Shape-of-Motion~\cite{som} & 15.38 & 0.483 & 0.396 \\
            MoSca~\cite{mosca} & 16.28 & 0.495 & 0.413 \\
            Ours & \textbf{17.69} & \textbf{0.556} & \textbf{0.292} \\
            \midrule
            MonST3R$^\dagger$~\cite{monst3r} & 9.32 & 0.103 & 0.710 \\
            4DGT$^\dagger$~\cite{4dgt} & 9.94 & 0.208 & 0.639 \\
            NeoVerse$^\dagger$~\cite{neoverse} & 11.56 & 0.293 & 0.558 \\
            Ours & \textbf{18.09} & \textbf{0.554} & \textbf{0.309} \\
            \bottomrule
        \end{tabular}
    \end{minipage}

    \vspace{0.8em}
    \begin{minipage}[t]{0.485\linewidth}
        \vspace{0pt}
        \centering
        \captionof{table}{\textbf{Results on MultiCamVideo~\cite{rcm}.}}
        \label{tab:dataset_results}
        \vspace{0.2em}
        \begin{tabular}{@{}lccc@{}}
            \toprule
            Method & \metricup{PSNR} & \metricup{SSIM} & \metricdown{LPIPS} \\
            \midrule
            Shape-of-Motion~\cite{som} & 21.17 & 0.750 & 0.245 \\
            MoSca~\cite{mosca} & 21.88 & 0.736 & 0.256 \\
            WorldTree~\cite{worldtree} & 20.76 & 0.792 & 0.285 \\
            ReCamMaster~\cite{rcm} & 19.48 & 0.596 & 0.255 \\
            ViDAR~\cite{vidar} & 22.14 & 0.754 & 0.198 \\
            Ours (sample) & \textbf{27.43} & \textbf{0.918} & \textbf{0.083} \\
            Ours (distill) & 24.36 & 0.855 & 0.118 \\
            \bottomrule
        \end{tabular}
    \end{minipage}\hfill
    \begin{minipage}[t]{0.485\linewidth}
        \vspace{0pt}
        \centering
        \captionof{table}{\textbf{Results on NVIDIA-LS~\cite{nvidials}.} Baselines from~\cite{worldtree}.}
        \label{tab:nvidia_ls}
        \vspace{0.2em}
        \begin{tabular}{@{}lccc@{}}
            \toprule
            Method & \metricup{PSNR} & \metricup{SSIM} & \metricdown{LPIPS} \\
            \midrule
            4DGS~\cite{wu20244dgs} & 23.34 & 0.790 & 0.128 \\
            MoSca~\cite{mosca} & 23.72 & 0.792 & 0.109 \\
            WorldTree~\cite{worldtree} & 24.06 & 0.804 & 0.100 \\
            Ours (distill) & \textbf{24.49} & \textbf{0.807} & \textbf{0.090} \\
            \bottomrule
        \end{tabular}
    \end{minipage}
\end{table}

\paragraph{In-the-wild data.}
We finally show qualitative results of our method on challenging in-the-wild real videos and those generated by an external method~\cite{veo}. ~\cref{fig:wild} shows our method compared to the input reconstruction~\cite{mosca}. The first sample is taken from a dynamic SLAM dataset WildGS~\cite{Zheng2025WildGS} and the second example is a generated video by~\cite{veo}. Notably, our model can successfully perform visual outpainting, fix highly-degraded dynamics, and even infer out-of-frustum dynamics while respecting the reconstruction of the static region. Please see the supplement for more video results.

\begin{figure}[t]
    \centering
    \begingroup
    \newcommand{\ablmethod}[1]{\makebox[0.166666\linewidth][c]{\scriptsize #1}}
    \ablmethod{No reference G-Buffer}%
    \ablmethod{No joint alignment}%
    \ablmethod{With 1.3B model}%
    \ablmethod{No 3D Scene Flow}%
    \ablmethod{Ours (Sample)}%
    \ablmethod{Ground Truth}%
    \par\vspace{0.2em}
    \includegraphics[width=\linewidth]{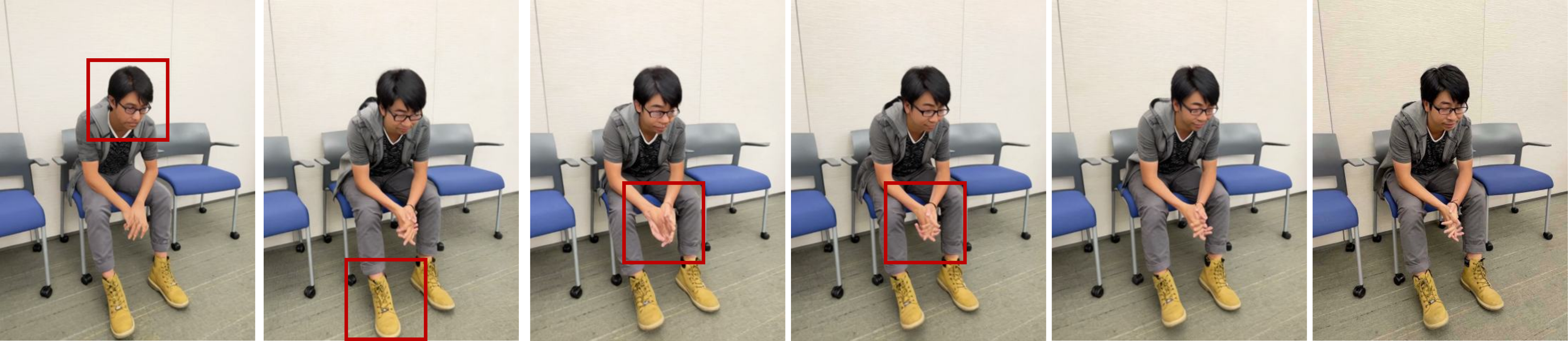}

    \vspace{0.6em}
    \ablmethod{No Motion Opt.}%
    \ablmethod{No reinitialization}%
    \ablmethod{No sample weight}%
    \ablmethod{One Virtual Camera}%
    \ablmethod{Ours (Distill)}%
    \ablmethod{Ground Truth}%
    \par\vspace{0.2em}
    \includegraphics[width=\linewidth]{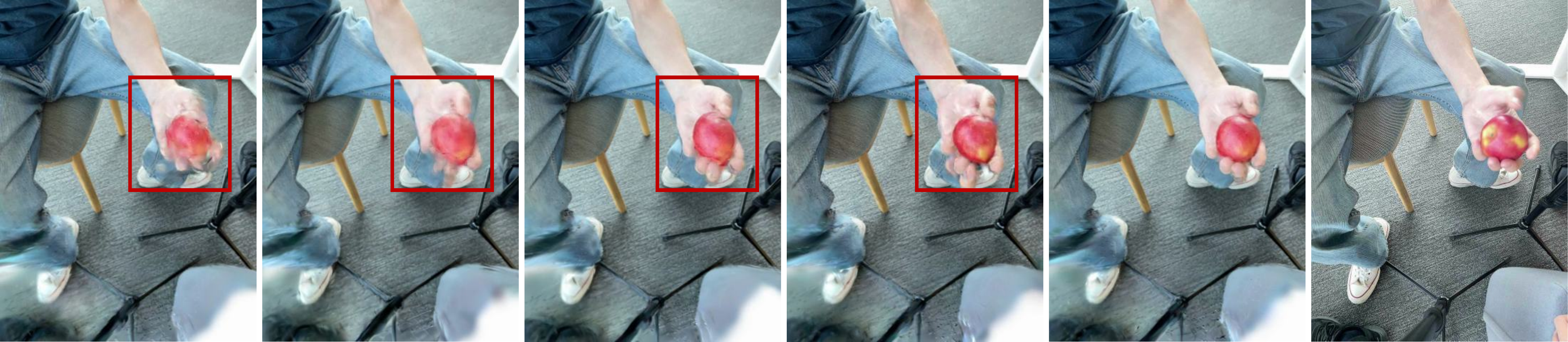}
    \endgroup
    \caption{\textbf{Ablation study of video model components (top) and re-optimization process (bottom).}}
    \label{fig:visual_ablations}
\end{figure}

\vspace{-0.5em}
\paragraph{Ablation of dynamic 3DGS-conditioned video model.}
In~\cref{tab:ablation} and the top row of~\cref{fig:visual_ablations}, we present ablations on our video model design and data pipeline. The first three rows of the second block highlight our conditioning signals. Removing 3D scene flow degrades motion coherence, removing camera injection for the input and target trajectories costs over 0.5\,dB, while removing reference-view G-buffers significantly degrades performance because the model loses reliable input-trajectory context under imperfect 4D reconstruction. We next examine the data pipeline. Joint depth alignment across multiview videos substantially outperforms independent per-video depth prediction because it preserves geometric consistency between the reconstructed 4D representation and target views. We also find that mitigating dataset bias~\cite{rcm} through temporal trimming and reversal is important because it prevents overfitting to fixed starting frames. Finally, we study model scaling by replacing the Wan 2.1 14B model with its 1.3B variant~\cite{wanvideo}. The resulting performance drop indicates that our approach benefits from larger-capacity video generators. Please see the supplement for more visual ablation comparisons.

\vspace{-0.5em}
\paragraph{Ablation of dynamic 3DGS refinement.}

In~\cref{tab:opt_ablation} and the bottom of~\cref{fig:visual_ablations}, we ablate key choices in our dynamic 3DGS refinement stage.
Fixing the motion scaffold (\textit{No motion opt.}) degrades performance, showing that re-optimizing motion is necessary to incorporate constraints from virtual views. Directly continuing optimization from the monocular reconstruction without reinitialization (\textit{No reinit.}) also leads to worse results, as the initial representation often overfits ill-posed single-view observations. Removing depth-based reinitialization (\textit{No depth reinit.}), where we augment the scene with multi-view depth from generated samples, reduces performance, indicating its importance for recovering missing regions.
Finally, treating generated views with equal importance as input views (\textit{No sample weight}) slightly hurts performance, suggesting that down-weighting synthesized observations improves robustness. We further evaluate the number of sampled virtual cameras in~\cref{tab:iphone_multiview_camera_count}. Reconstruction quality improves monotonically with additional virtual viewpoints, with gains largely saturating beyond eight cameras.

\begin{table}[t]
\begin{minipage}[t]{0.48\columnwidth}
    \vspace{0pt}
    \centering
    \resulttablesize
    \captionof{table}{\textbf{Ablation study of video model.}}
    \label{tab:ablation}

    \vspace{0.15em}
    \setlength{\tabcolsep}{2pt}
    \resizebox{0.9\linewidth}{!}{%
    \begin{tabular}{@{}lccc@{}}
        \toprule
        Variant & \metricup{PSNR} & \metricup{SSIM} & \metricdown{LPIPS} \\
        \midrule
        Ours & \textbf{19.50} & \textbf{0.701} & 0.224 \\
        No Recon. Guidance & 19.43 & 0.690 & \textbf{0.223} \\
        \midrule
        No 3D scene flow & 19.30 & 0.688 & 0.226 \\
        No camera injection & 18.93 & 0.693 & 0.229 \\
        No reference G-Buffer & 18.61 & 0.678& 0.265\\
        No joint alignment & 15.58 & 0.598 & 0.347 \\
        No debiased data & 18.57 & 0.665 & 0.250 \\
        With 1.3B model & 18.64 & 0.671 & 0.244 \\
        \bottomrule
    \end{tabular}%
    }
\end{minipage}\hfill
\begin{minipage}[t]{0.45\columnwidth}
    \vspace{0pt}
    \centering
    \resulttablesize
    \captionof{table}{\textbf{Ablation study of scene refinement.}}
    \label{tab:opt_ablation}

    \vspace{0.15em}
    \setlength{\tabcolsep}{2pt}
    \resizebox{0.9\linewidth}{!}{%
    \begin{tabular}{@{}lccc@{}}
        \toprule
        Variant & \metricup{PSNR} & \metricup{SSIM} & \metricdown{LPIPS} \\
        \midrule
        Ours & \textbf{19.49} & 0.704 & \textbf{0.177} \\
        No motion opt. & 18.64 & 0.680 & 0.209 \\
        No reinit. & 19.15 & 0.695 & 0.218 \\
        No depth reinit. & 19.40 & \textbf{0.705} & 0.195 \\
        No sample weight & 19.36 & 0.700 & 0.190 \\
        \bottomrule
    \end{tabular}%
    }
\end{minipage}
\end{table}

\section{Discussion}
We have presented \method, a 4D reconstruction framework for real-world monocular videos. By conditioning on reconstructed dynamic 3DGS, we show how to attain highly precise and physically plausible video samples and how to re-optimize those samples into a higher-quality dynamic 3DGS. Our approach significantly reduces the burden of high-quality 4D capture. However, it has some limitations. When the initial scene reconstruction fails completely, our method struggles to recover. The video model may also struggle to align the reconstruction with the input when the target trajectory moves extremely far from the input trajectory. Please see the supplement for representative failure cases.

We believe there are many promising future directions for our method. Incorporating multiple rounds of alternation between refinement and sampling of the dynamic 3DGS reconstruction may produce higher-quality results. We also hope that our method may be useful in downstream robotic applications, reducing the camera burden and improving planning. Finally, we hope that our approach can form a basis for memory and persistence in video generative modeling.

\section*{Acknowledgments}
We thank Yang Zheng, Zhengfei Kuang, Lior Yariv, and Jianhao Zheng for fruitful discussions. We also thank Yijia Weng and Jiahui Lei for providing evaluation details for MoSca, Kuan Heng Lin for providing Vista4D evaluation details, and Michal Nazarczuk and Eduardo P{\'e}rez-Pellitero for providing evaluation details for ViDAR.

\newpage
\bibliographystyle{plainnat}
\bibliography{references}
\clearpage
\appendix
\clearpage
\begin{center}
    {\Large \textbf{Supplementary Material for World from Motion}}
\end{center}
\vspace{1em}

\setcounter{section}{0}
\setcounter{figure}{0}
\setcounter{table}{0}
\renewcommand{\thetable}{S\arabic{table}}
\renewcommand{\thefigure}{S\arabic{figure}}
\renewcommand{\thesection}{S\arabic{section}}

\begin{abstract}
    This supplementary document provides additional technical details, experimental analyses, and qualitative results to complement the main paper. We present a detailed analysis of test-time pose optimization for consistent comparison, comprehensive implementation and evaluation protocols, and a granular breakdown of performance across dynamic and static scene components. Furthermore, we include additional qualitative ablations and a discussion of representative failure cases to characterize the current limitations of our framework.
\end{abstract}

\section{Pose optimization analysis.}\label{sec:pose_optimization}
\begin{figure}[ht]
    \centering
    \captionsetup[subfigure]{labelformat=empty}
    \renewcommand\thesubfigure{\thefigure\alph{subfigure}}
    \makeatletter
    \renewcommand\p@subfigure{}
    \makeatother
    \caption{\textbf{Pose optimization analysis.} Left: accumulated optical flow of the static test camera with and without test-time per-frame optimization. Right: test camera trajectories under different optimization strategies.}
    \label{fig:pose_optimization}

    \vspace{0.35em}
    \newsavebox{\poseheaderbox}
    \sbox{\poseheaderbox}{\scriptsize Per-Timestep Pose}
    \begin{subfigure}[t]{0.55\columnwidth}
        \centering
        \setlength{\tabcolsep}{0pt}
        \newsavebox{\posecomparisonbox}
        \sbox{\posecomparisonbox}{%
            \includegraphics[width=0.84\linewidth,height=0.34\textheight,keepaspectratio]{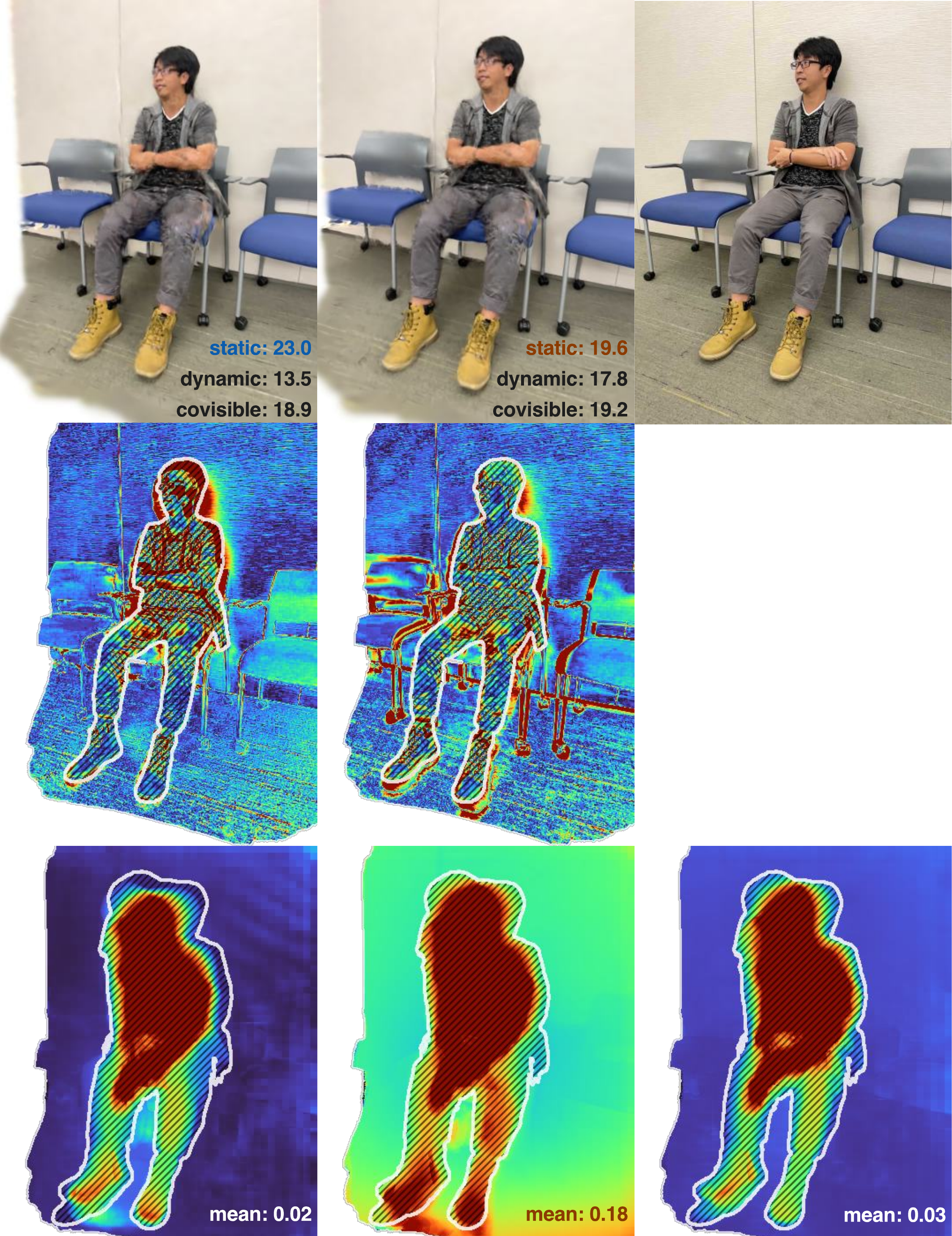}%
        }
        \newlength{\posecomparisonheight}
        \setlength{\posecomparisonheight}{\dimexpr\ht\posecomparisonbox+\dp\posecomparisonbox\relax}
        {\scriptsize
        \begin{tabular}[t]{@{}c@{\hspace{0.2em}}c@{}}
            &
            \begin{minipage}[t]{\wd\posecomparisonbox}
                \vspace{0pt}
                \makebox[0.333\linewidth]{Corrected Pose}%
                \makebox[0.333\linewidth]{Per-Timestep Pose}%
                \makebox[0.333\linewidth]{Ground Truth}%
            \end{minipage}
            \\[-0.1em]
            \begin{minipage}[t][\posecomparisonheight][t]{0.06\linewidth}
                \vspace{0pt}
                \parbox[c][\dimexpr\posecomparisonheight/3\relax][c]{\linewidth}{\centering\makebox[0pt][c]{\rotatebox{90}{RGB}}}\par
                \parbox[c][\dimexpr\posecomparisonheight/3\relax][c]{\linewidth}{\centering\makebox[0pt][c]{\rotatebox{90}{Error}}}\par
                \parbox[c][\dimexpr\posecomparisonheight/3\relax][c]{\linewidth}{\centering\makebox[0pt][c]{\rotatebox{90}{Aggregate Flow}}}%
            \end{minipage}
            &
            \begin{minipage}[t]{\wd\posecomparisonbox}
                \vspace{0pt}
                \usebox{\posecomparisonbox}
            \end{minipage}
        \end{tabular}}
        \phantomsubcaption
        \label{fig:pose_optimization_comparison}
    \end{subfigure}\hfill
    \begin{minipage}[t]{0.01\columnwidth}
        \centering
        \vspace{0pt}
        \smash{\raisebox{0pt}[0pt][0pt]{%
            \textcolor{black!25}{\rule[-0.41\textheight]{0.6pt}{0.41\textheight}}%
        }}
    \end{minipage}\hfill
    \begin{subfigure}[t]{0.415\columnwidth}
        \centering
        \vphantom{\usebox{\poseheaderbox}}\\[-0.1em]
        \includegraphics[width=\linewidth]{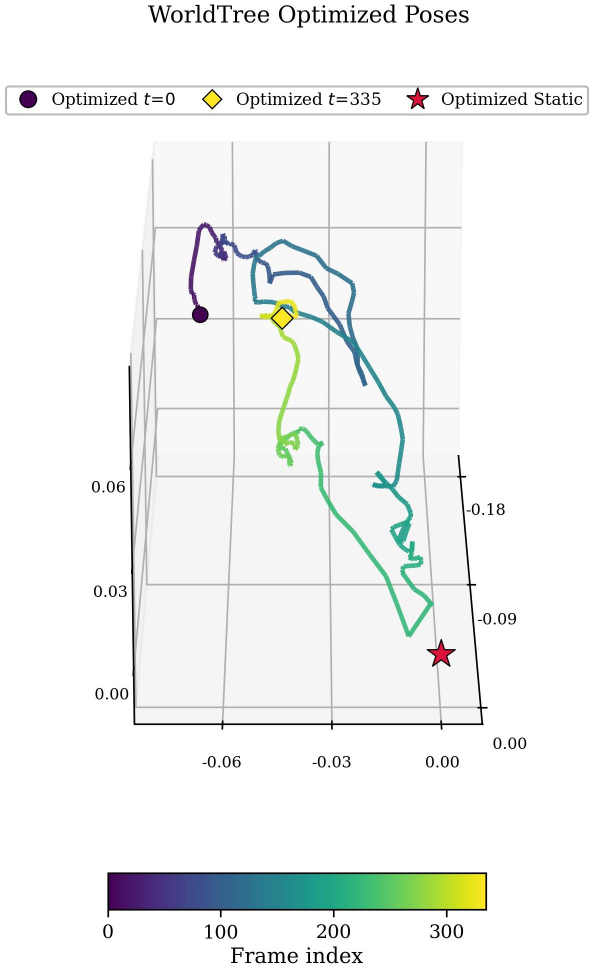}
        \phantomsubcaption
        \label{fig:optimized_poses}
    \end{subfigure}
\end{figure}

MoSca~\cite{mosca} and methods that build on it~\cite{vidar, worldtree} perform optimization through their output dynamic 3DGS to refine the test poses. They use both dynamic and static pixels to optimize a different camera pose per frame, as shown on the right of Figure~\ref{fig:optimized_poses}. However, the test camera poses of DyCheck are \emph{static}. These methods therefore trade off camera pose accuracy for better metrics in the dynamic region, as seen on the top left of Figure~\ref{fig:pose_optimization_comparison}. As seen in the middle left of Figure~\ref{fig:pose_optimization_comparison}, the metrics in the static region degrade significantly while the dynamic region improves, slightly improving the overall metrics. The aggregate flow of the test image should be approximately zero in the non-dynamic region outlined in white, as seen in the ground-truth aggregate flow (predicted by RAFT-large~\cite{raft}) on the bottom of Figure~\ref{fig:optimized_poses}, but this is violated in their per-timestep pose-optimization procedure.

To rectify this, we optimize one camera pose for the whole test sequence using only the static region. The comparison between the former pose optimization and our corrected version in ~\cref{tab:tto_masked_results} shows that this improves MoSca's performance in the static region, but decreases its overall metrics.

\vspace{0.8em}

\noindent
\begin{minipage}{\columnwidth}
    \centering
    \refstepcounter{table}\label{tab:tto_masked_results}
    \resulttablesize
    \textbf{Table \thetable}: MoSca results with our corrected pose
    optimization.

    \vspace{0.35em}
    \setlength{\tabcolsep}{1.6pt}
    \resizebox{0.9\columnwidth}{!}{%
        \begin{tabular}{@{}l*{3}{c}@{\hspace{3.5pt}}*{3}{c}@{\hspace{3.5pt}}*{3}{c}@{}}
            \toprule
            & \multicolumn{3}{c}{Covisible}
            & \multicolumn{3}{c}{Static Intersection}
            & \multicolumn{3}{c}{Dynamic Intersection} \\
            \cmidrule(lr){2-4}
            \cmidrule(lr){5-7}
            \cmidrule(lr){8-10}
            Pose Opt.
            & PSNR $\uparrow$ & SSIM $\uparrow$ & LPIPS $\downarrow$
            & PSNR $\uparrow$ & SSIM $\uparrow$ & LPIPS $\downarrow$
            & PSNR $\uparrow$ & SSIM $\uparrow$ & LPIPS $\downarrow$ \\
            \midrule
            All Pixels
            & \textbf{19.33} & \textbf{0.708} & \textbf{0.264}
            & 20.64 & 0.803 & 0.258
            & \textbf{15.93} & \textbf{0.909} & \textbf{0.242} \\
            Static Only
            & 18.69 & 0.696 & 0.272
            & \textbf{20.90} & \textbf{0.806} & \textbf{0.251}
            & 14.71 & 0.895 & 0.266 \\
            \bottomrule
        \end{tabular}%
    }
\end{minipage}

\vspace{0.8em}

\vspace{0.8em}

\begin{figure}[t]
    \centering
    \begingroup
    \newcommand{\vcammethod}[1]{\makebox[0.2\linewidth][c]{\scriptsize #1}}
    \vcammethod{0 Virtual Cameras (MoSca)}%
    \vcammethod{1 Virtual Camera}%
    \vcammethod{2 Virtual Cameras}%
    \vcammethod{8 Virtual Cameras}%
    \vcammethod{Ground Truth}%
    \par\vspace{0.2em}
    \includegraphics[width=\linewidth]{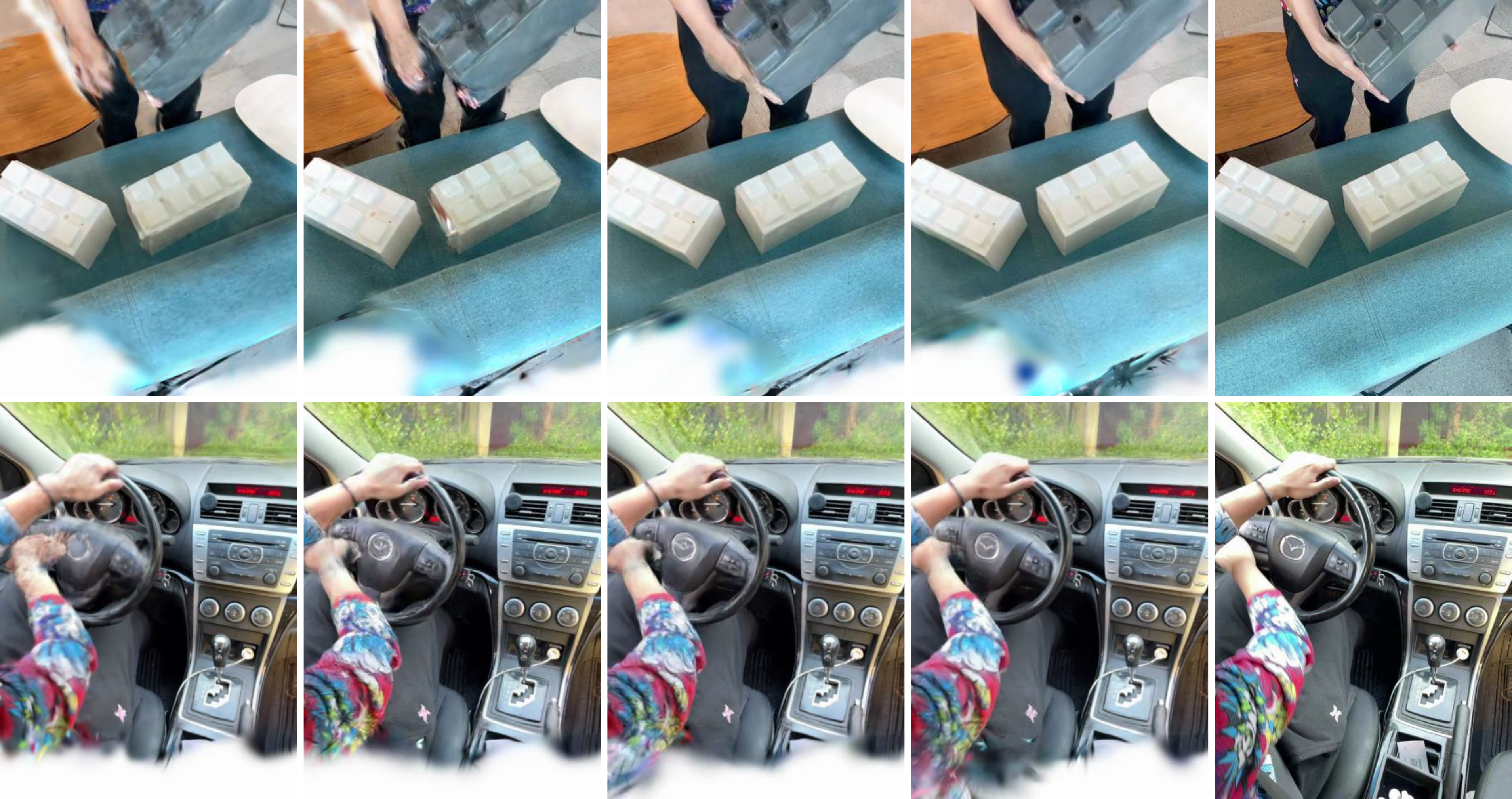}
    \endgroup
    \caption{\textbf{Qualitative comparison on DyCheck as the number of virtual
    cameras varies.} MoSca corresponds to the no-virtual-camera setting and
    suffers the most degradation; quality improves as more virtual cameras
    are added.}
    \label{fig:virtual_camera_qualitative}
\end{figure}

\paragraph{Virtual camera count qualitative comparison.}
Figure~\ref{fig:virtual_camera_qualitative} provides a qualitative companion to
\cref{tab:iphone_multiview_camera_count}, showing held-out target views on DyCheck as the number of synthesized virtual cameras varies. MoSca, which
uses no virtual cameras, exhibits the most artifacts; reconstruction quality
improves steadily as additional virtual cameras are added.

\vspace{0.8em}

\begin{figure}[t]
    \centering
    \begingroup
    \newcommand{\synmethod}[1]{\makebox[0.25\linewidth][c]{\scriptsize #1}}
    \synmethod{MoSca}%
    \synmethod{ViDAR}%
    \synmethod{Ours (Sample)}%
    \synmethod{Ground Truth}%
    \par\vspace{0.2em}
    \includegraphics[width=0.499\linewidth,trim={0 0 1944pt 0},clip]{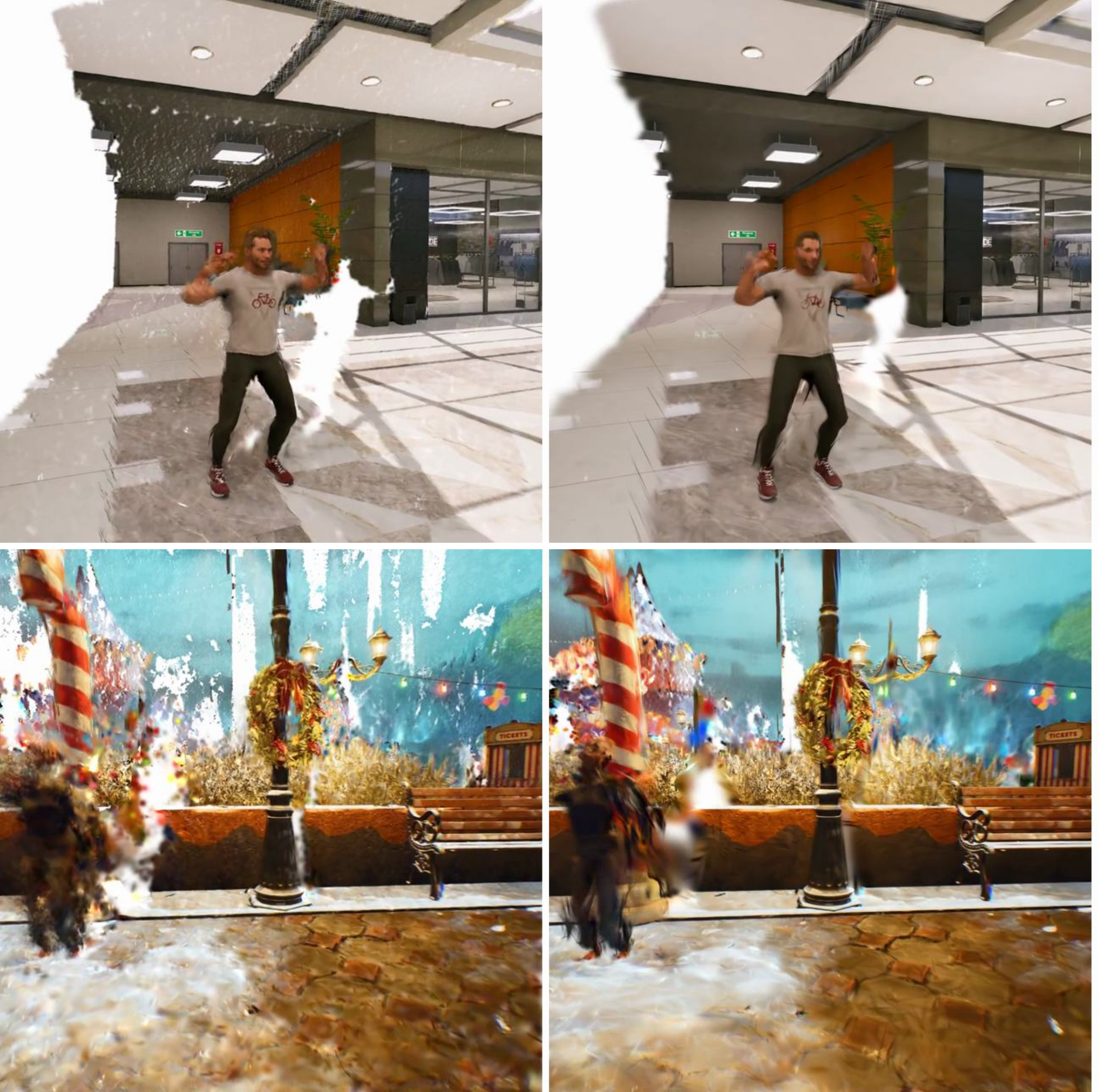}\hspace{0.002\linewidth}%
    \includegraphics[width=0.499\linewidth,trim={1944pt 0 0 0},clip]{figures/synthetic_qual.pdf}
    \endgroup
    \caption{\textbf{Qualitative comparison on the MultiCamVideo~\cite{rcm} benchmark.} Each row shows a held-out novel view; our method produces sharper details and more coherent motion than baseline reconstructions.}
    \label{fig:qualitative_synthetic}
\end{figure}

\paragraph{MultiCamVideo qualitative comparison.}
Figure~\ref{fig:qualitative_synthetic} provides a qualitative companion to~\cref{tab:dataset_results}, showing held-out novel-view renderings on our MultiCamVideo~\cite{rcm} validation split. Compared to MoSca~\cite{mosca} and ViDAR~\cite{vidar}, our samples recover sharper details and reconstruct more coherent motion under large viewpoint changes.

\paragraph{Dynamic and static valid breakdown.}
We further analyze our model's performance by decomposing the improvements across the dynamic and static components of the scene. To do so, we obtain dynamic segmentation masks for all test images using SAM3~\cite{sam3}. Metrics are then computed over the intersection of the valid evaluation region with the static and dynamic masks, respectively. As shown in Table~\ref{tab:results_valid_parts}, our method consistently improves rendering quality across both static and dynamic regions, demonstrating robust refinement capabilities regardless of scene motion.

\noindent
\begin{minipage}{\columnwidth}
    \refstepcounter{table}
    \centering
    \resulttablesize
    \textbf{Table \thetable}: \textbf{Dynamic and static valid metrics for the same
    comparison methods.}
    \label{tab:results_valid_parts}

    \vspace{0.35em}
    \setlength{\tabcolsep}{3pt}
    \begin{tabular}{@{}l*{3}{c}@{\hspace{6pt}}*{3}{c}@{}}
        \toprule
        & \multicolumn{3}{c}{Dynamic Valid}
        & \multicolumn{3}{c}{Static Valid} \\
        \cmidrule(lr){2-4}
        \cmidrule(lr){5-7}
        Method
        & \metricup{mPSNR} & \metricup{mSSIM} & \metricdown{mLPIPS}
        & \metricup{mPSNR} & \metricup{mSSIM} & \metricdown{mLPIPS} \\
        \midrule
        MoSca & 14.58 & 0.881 & 0.303 & 19.52 & 0.775 & 0.271 \\
        WT & 15.15 & 0.887 & 0.315 & 20.00 & 0.786 & 0.227 \\
        ViDAR & 15.79 & 0.893 & 0.280 & 20.10 & 0.780 & 0.218 \\
        \midrule
        Ours+MS & \textbf{15.84} & \textbf{0.894} & \textbf{0.191} & \textbf{20.56} & 0.779 & \textbf{0.194} \\
        Ours+WT & 15.69 & 0.891 & 0.264 & 20.16 & \textbf{0.787} & 0.222 \\
        \bottomrule
    \end{tabular}
\end{minipage}

\vspace{0.8em}

\begin{figure}[t]
    \centering
    \begin{minipage}{0.5\columnwidth}
        \centering
        \begingroup
        \setlength{\tabcolsep}{0pt}
        \scriptsize
        \begin{tabular}{@{}cc@{}}
            \makebox[0.5\linewidth][c]{Reference Image} &
            \makebox[0.5\linewidth][c]{Masked Test Image} \\
        \end{tabular}
        \par\vspace{0.2em}
        \includegraphics[width=\linewidth]{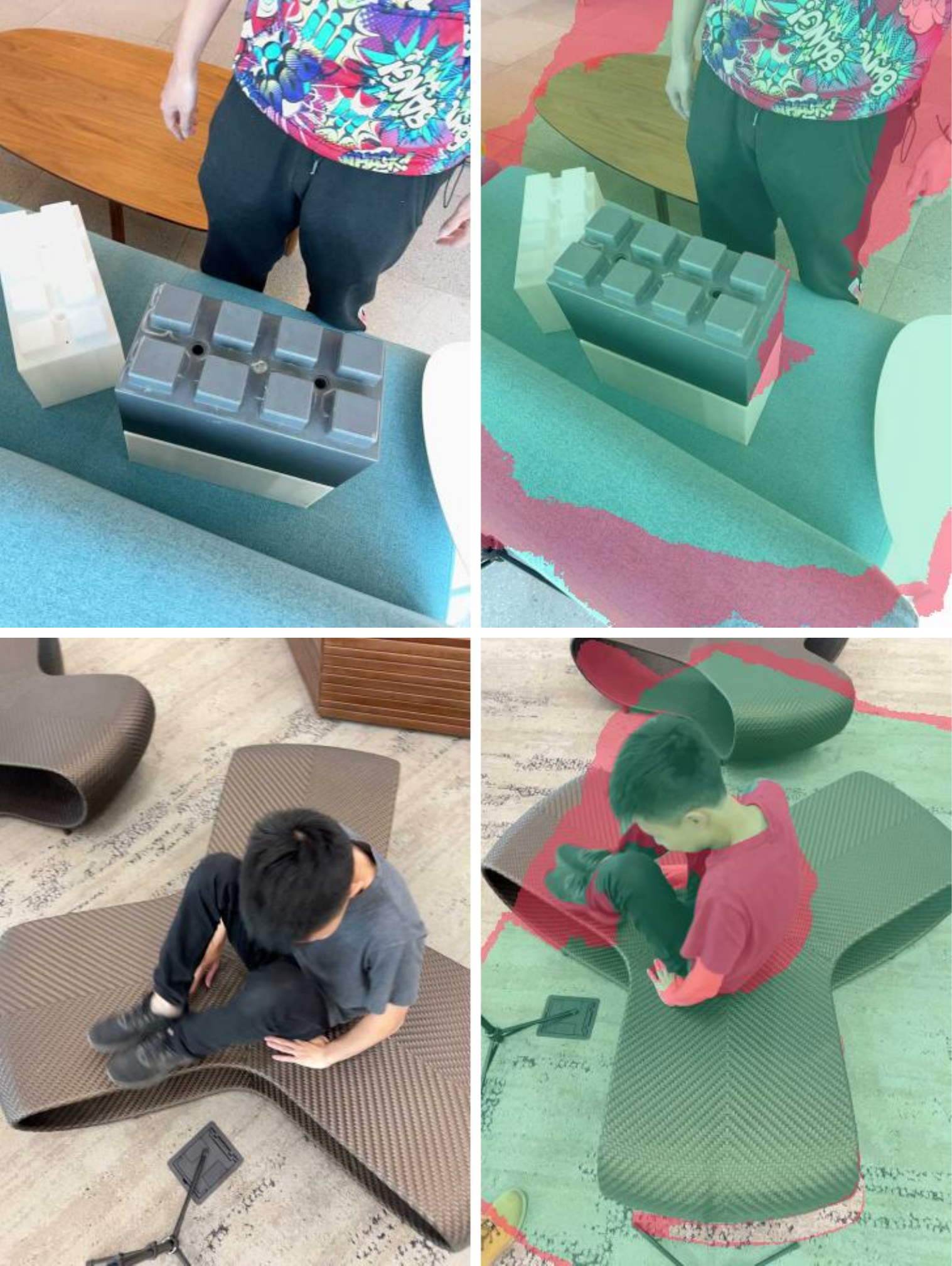}
        \endgroup
        \caption{\textbf{Evaluation mask visualization.} Reference image and corresponding masked test image used
        for evaluation on DyCheck.}
        \label{fig:mask_viz}
    \end{minipage}
\end{figure}

\paragraph{Test mask visualization.}
Figure~\ref{fig:mask_viz} shows a representative DyCheck test pair: the reference image used as model input alongside the corresponding masked test image. We define specific evaluation regions to isolate different aspects of reconstruction quality: the green mask identifies instantaneous covisible pixels between the input and target viewpoints, while the union of the red and green regions constitutes the valid region. This latter region is used to evaluate the fidelity of spatiotemporal accumulation within a persistent 4D reconstruction.

\begin{figure}[!htbp]
    \centering
    \includegraphics[width=0.92\columnwidth,trim={110pt 130pt 105pt 125pt},clip]{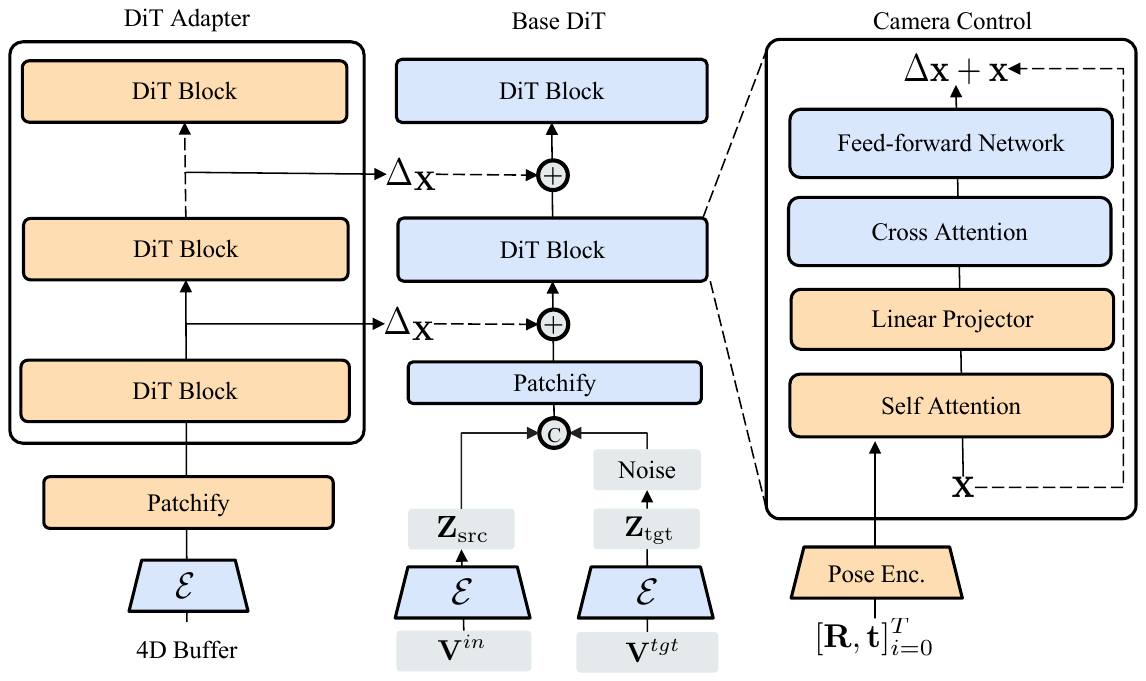}
    \caption{\textbf{Network architecture.} Our generator combines a
    4D-buffer DiT adapter with a pretrained video DiT. The adapter encodes
    rendered 4D buffers and injects residual updates into the base DiT, while a
    camera-control module conditions self-attention on the relative input and
    target camera trajectories.}
    \label{fig:network_architecture}
\end{figure}

\clearpage
\paragraph{Network architecture.}
Figure~\ref{fig:network_architecture} illustrates the architecture of our
4D-conditioned video generator. The model builds on a video DiT backbone that
denoises the concatenation of clean input-video latents and noisy target-video
latents, with the flow-matching loss applied only to the target branch. In
parallel, a VACE-style DiT adapter processes the rendered 4D buffers from the
input and target trajectories. Its residual features are injected into the base
DiT blocks, allowing the generator to use the initial dynamic 3DGS as a dense
spatiotemporal scaffold while still relying on the pretrained video prior for
photorealistic synthesis. Camera control is added through a pose encoder that
maps relative camera trajectories to token-level conditioning before
self-attention, giving the denoising network explicit geometric information for
both observed and target views.

\paragraph{Implementation details.}
We employ Wan-2.1-14B T2V~\cite{wanvideo} as the backbone for our dynamic 3DGS-conditioned video generator. The DiT adapter is initialized from pretrained Wan-14B-2.1-VACE weights~\cite{vace}. VACE's context embedding takes 96 latent channels; our five buffer modalities give $5\times16=80$, so we replace its patch-embedding layer with a new 3D convolution taking 80 channels. During training, we fine-tune the full DiT adapter blocks, the camera encoder, and the self-attention layers within the base DiT, while keeping all other parameters frozen.
The training data consists of 150K 73-frame video pairs at a resolution of 512$\times$384 derived from SoM~\cite{som} and WorldTree~\cite{worldtree} reconstructions. We set the learning rate to $10^{-5}$ and the batch size to 64, and train our model for 12K steps using the AdamW optimizer~\cite{adamw}. At inference time, we set the number of denoising steps to 20 for all experiments.
For the re-optimization stage, we set $\lambda_{\mathrm{tgt}}=0.5$, $\lambda_{\mathrm{motion}}=3.0$, $\lambda_{\mathrm{depth}}=0.05$, $\lambda_{\mathrm{track}}=0.01$, and $\lambda_{\mathrm{arap}}=3.0$. We run re-optimization for 15K steps in all experiments.

\paragraph{Dataset debiasing.}\label{sec:debiasing}
In MultiCamVideo~\cite{rcm}, all videos share the same starting viewpoint, introducing a bias uncommon in real-world capture. To address this, we apply temporal augmentations during training, including random temporal cropping and sequence reversal. This exposes the model to diverse trajectories and temporal orders, improving robustness to varied camera trajectories.

\paragraph{Evaluation details.}
For the MultiCamVideo dataset, the original test set lacks ground-truth camera parameters. Since our task requires precise poses to evaluate 4D reconstruction accuracy, we instead hold out a custom test split from the source data. This split consists of 5 scenes; for each, we randomly select one camera as the input view and use the remaining 5 cameras for evaluation. We evaluate all methods on the first 73 frames at a spatial resolution of 640$\times$640. On DyCheck, we run the initial reconstruction at a resolution of 480$\times$360. For initial 4D reconstruction, we use RAFT~\cite{raft} for optical flow extraction and Tapir~\cite{doersch2023tapir} for 2D track extraction, following MoSca to ensure a consistent comparison across methods.
To handle sequences exceeding 300 frames, we adopt a temporal sliding window approach to process the videos in chunks. Generation uses non-overlapping 73-frame windows (the training length), with the final window shifted back so that it also receives a full 73-frame context. Since every window is conditioned on buffers rendered from one global 4D reconstruction, windows cannot drift geometrically relative to each other, and residual per-window appearance variation is absorbed during distillation. For model inference, 4D buffers are resized to our training resolution of 512$\times$384, with the resulting videos resized back to the target resolution to ensure a consistent comparison with baseline methods. For in-the-wild samples, we apply a temporal stride of 2 to the input videos, allowing the model to incorporate longer temporal context.

\paragraph{Camera sampling strategy.} On DyCheck, we use our main method to sample the 8 farthest cameras along the input trajectory and fix them over time. In the MultiCamVideo~\cite{rcm} dataset, each scene contains ten camera trajectories; we randomly select one of these as the virtual camera for dynamic 3DGS re-optimization. For in-the-wild samples, we adopt the camera UI from \cite{vista4d} and manually sample novel trajectories to serve as virtual cameras.

\paragraph{Normal rendering.} We derive normals from a rendered world-position map $\mathbf{P}_t=\sum_i \alpha_i T_i \boldsymbol{\mu}_i(t)$, obtained by alpha-compositing Gaussian centers, via central differences, $\mathbf{N}_t=\mathrm{normalize}(\partial_x\mathbf{P}_t\times\partial_y\mathbf{P}_t)$, oriented toward the camera and zeroed at silhouettes and depth discontinuities. Normals are stored in world space, so they remain consistent as the camera moves.

\begin{figure}[t]
   \centering
   \includegraphics[width=\columnwidth]{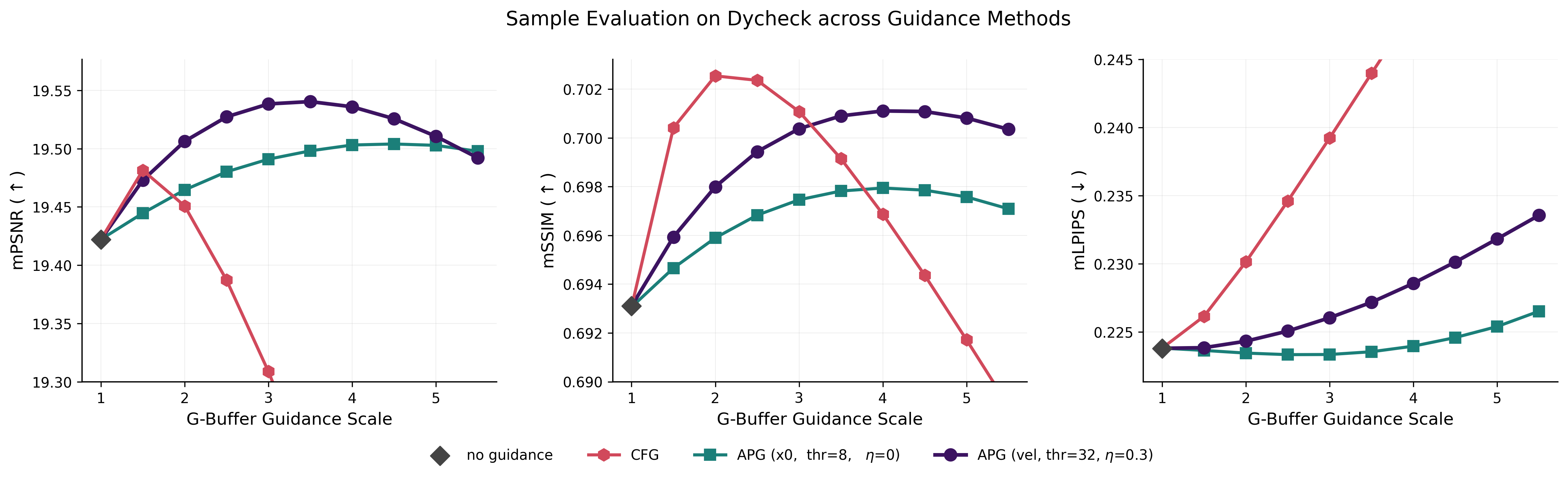}
   \caption{Guidance-scale sweep on DyCheck across no guidance, CFG, and two
   APG variants. Velocity APG is the most stable at higher scales and reaches
   the strongest mPSNR, while the $x_0$ APG variant yields the best
   mLPIPS.}
   \label{fig:guidance_comparison_dycheck}
\end{figure}

\paragraph{Reconstruction guidance ablation.}
Figure~\ref{fig:guidance_comparison_dycheck} sweeps the guidance scale on DyCheck for no guidance, standard classifier-free guidance (CFG)~\cite{cfg}, and two APG~\cite{apg}
variants. This complements Table~\ref{tab:ablation}: velocity APG gives the
strongest mPSNR/mSSIM trade-off near the selected operating point, while the
$x_0$ APG variant achieves the lowest mLPIPS.

\begin{figure}[t]
   \centering
   {\small
   \begin{tabular}{@{}*{4}{c}@{}}
       \makebox[0.22\columnwidth]{No Guidance} &
       \makebox[0.22\columnwidth]{CFG=1.5} &
       \makebox[0.22\columnwidth]{APG=3.5} &
       \makebox[0.22\columnwidth]{Ground Truth}
   \end{tabular}
   \par\vspace{0.2em}}
   \includegraphics[width=\columnwidth]{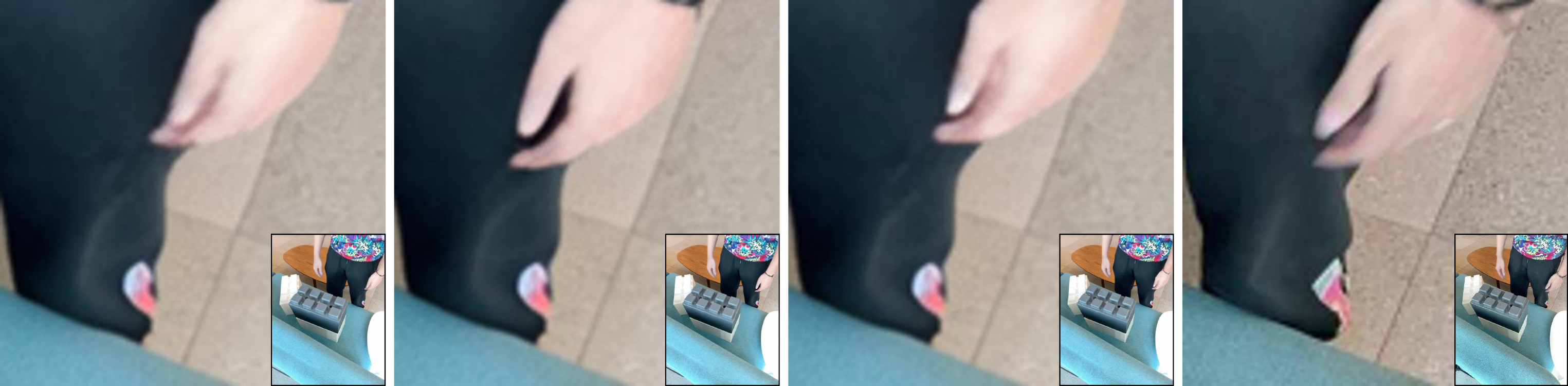}
   \caption{Qualitative guidance comparison for representative DyCheck outputs.}
  \label{fig:guidance_comparison_examples}
\end{figure}

\paragraph{Generalization.}
To demonstrate the robustness of our design across different 4D representations, we evaluate a version of our model trained exclusively on SoM-derived data and test it using DyCheck reconstructed by MoSca~\cite{mosca} and WorldTree~\cite{worldtree}. We report the performance of direct samples at test views with full image metrics. Table~\ref{tab:generalization} shows each baseline on
its own row, followed by a `+ Ours' row whose green parenthesized values denote
the change relative to the baseline directly above. Our model trained on a different, more degraded representation can still faithfully improve the reconstruction fidelity (PSNR) and perceptual quality (LPIPS), with SSIM slightly falling behind.

\noindent
\begin{minipage}{\columnwidth}
    \refstepcounter{table}
    \centering
    \resulttablesize
    \textbf{Table \thetable}: Our method trained on Shape of Motion~\cite{som} generalizes to improve other methods.
    \label{tab:generalization}

    \vspace{0.35em}
    \setlength{\tabcolsep}{1.5pt}
    \begin{tabular}{@{}lccccc c@{}}
        \toprule
        Method
        & \multicolumn{2}{c}{\metricup{PSNR}}
        & \multicolumn{2}{c}{\metricup{SSIM}}
        & \multicolumn{2}{c}{\metricdown{LPIPS}} \\
        \cmidrule(lr){2-3}
        \cmidrule(lr){4-5}
        \cmidrule(lr){6-7}
        MoSca~\cite{mosca}\textsuperscript{*} & 16.60 & & 0.553 & & 0.362 & \\
        + Ours & 18.26 & \deltagreen{+1.66} & 0.540 & \deltared{-0.013} & 0.278 & \deltagreen{-0.084} \\
        \midrule
        WT~\cite{worldtree}\textsuperscript{*} & 17.08 & & 0.574 & & 0.332 & \\
        + Ours & 18.60 & \deltagreen{+1.52} & 0.550 & \deltared{-0.024} & 0.274 & \deltagreen{-0.058} \\
        \bottomrule
    \end{tabular}
\end{minipage}

\vspace{0.8em}

\begin{figure}[t]
    \centering
    \begin{minipage}{0.5\columnwidth}
        \centering
        \begingroup
        \newcommand{\failmethod}[1]{\makebox[0.333\linewidth][c]{\scriptsize #1}}
        \failmethod{Input Frame}%
        \failmethod{Reconstruction}%
        \failmethod{Output}%
        \par\vspace{0.2em}
        \includegraphics[width=\linewidth]{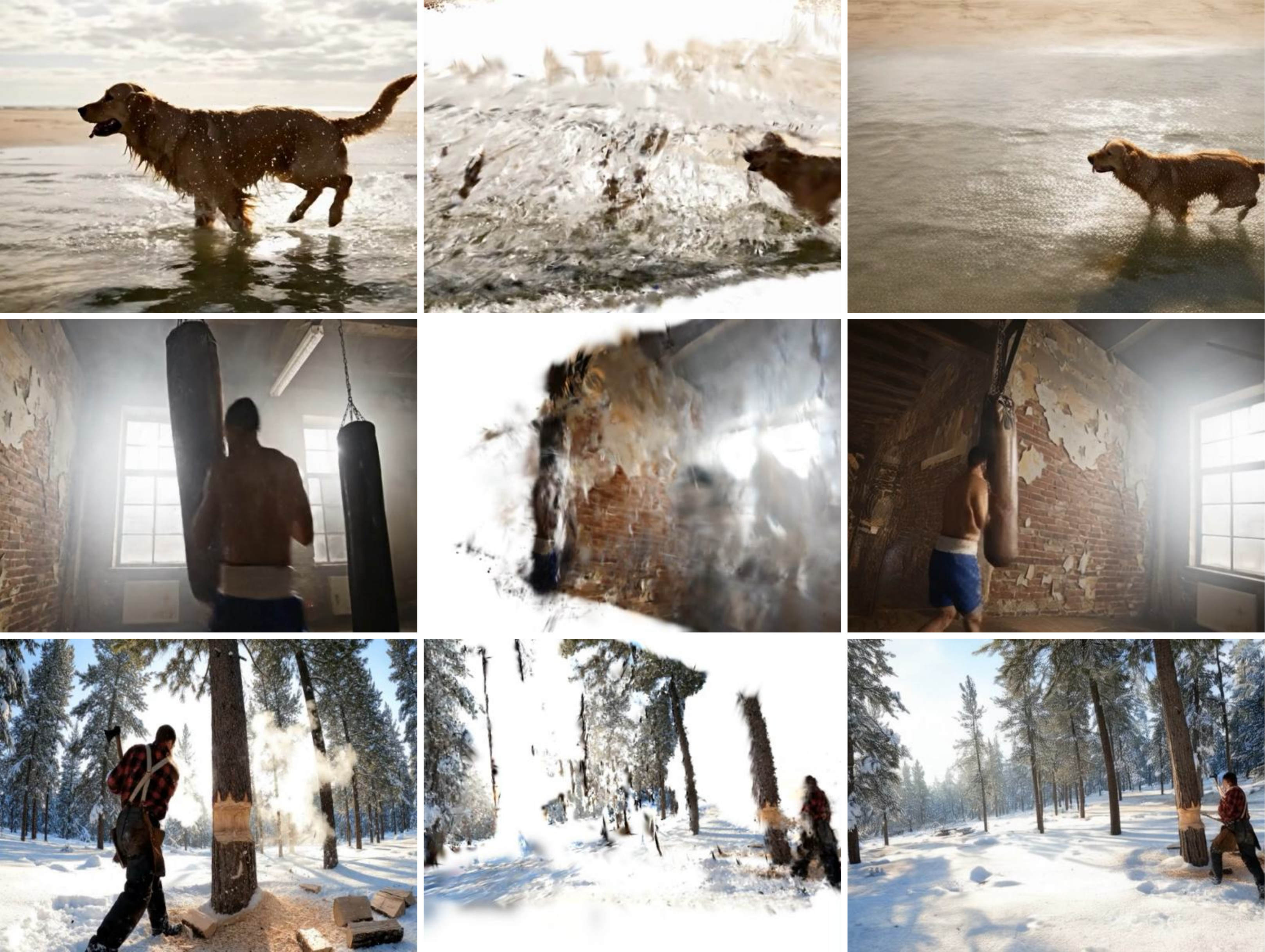}
        \endgroup
        \caption{\textbf{Representative failure cases of \method.}}
        \label{fig:failures}
    \end{minipage}
\end{figure}

\paragraph{Failure cases.}
Figure~\ref{fig:failures} illustrates representative failure cases of \method. We observe that artifacts present in the initial input reconstruction can occasionally propagate to the final output despite our generative refinement process. Specifically, the model struggles with scenes containing complex stochastic or volumetric effects, such as the fluids and smoke shown in the first and third examples. Additionally, as seen in the second example, our model may fail to accurately hallucinate consistent dynamics when subjected to extreme viewpoint shifts.

\paragraph{Compute resources.}
Our conditional video generative model was trained on a cluster of 32 NVIDIA Blackwell GB200 GPUs (192 GB VRAM each). The final model reached convergence in approximately 48 hours of training. For the dynamic 3DGS re-optimization stage, all experiments were conducted on NVIDIA A100 (80 GB) GPUs. On average, the optimization of a 400-frame sequence requires 40 minutes to complete on a single A100.

\paragraph{Societal impacts.} Our approach unifies geometric reconstruction with temporal priors from generative models. Its deployment and the advancement of 4D world modeling carry both positive and negative impacts.
On the positive side, by providing more consistent and precise 4D reconstruction from monocular video, our work can enhance the "world models" used by autonomous systems. This has the potential to improve efficiency and safety in human-robot interaction and navigation in complex environments. Our method also lowers the barrier for creators in AR/VR and education, allowing for the preservation of dynamic heritage or personal moments in 4D.
On the negative side, the generative capabilities of \method present potential for misuse in the creation of deceptive 4D content. While our method is designed for scene reconstruction, the underlying temporal priors could be used to generate fraudulent "deepfake" videos that exhibit high spatiotemporal consistency. Additionally, the significant energy consumption required to train 14B-parameter generative backbones poses environmental challenges.

\paragraph{Safeguards.}
To ensure responsible use, we plan to release our code and model weights under a standard academic license that restricts misuse.

\paragraph{Data licenses.} MultiCamVideo~\cite{rcm}, DyCheck~\cite{dycheck}, and WildGS~\cite{Zheng2025WildGS} are under Apache 2.0.

\end{document}